\documentclass[10pt,twocolumn,letterpaper]{article}

\usepackage[final]{cvpr}
\usepackage[pagebackref,breaklinks,colorlinks]{hyperref}
\usepackage{times}
\usepackage{epsfig}
\usepackage{graphicx}
\usepackage{amsmath}
\usepackage{amssymb}
\usepackage{times}
\usepackage{soul}
\usepackage{url}
\usepackage[utf8]{inputenc}
\usepackage{amsmath}
\usepackage{amsthm}
\usepackage{booktabs}
\usepackage{algorithm}
\usepackage{algorithmic}
\usepackage{amsmath}
\usepackage{amssymb}
\usepackage{booktabs}
\usepackage{multirow}
\usepackage{paralist,algorithmic,algorithm}
\usepackage[american]{babel}
\usepackage{microtype}
\usepackage{lipsum}
\usepackage{bm}
\usepackage[switch]{lineno}  %
\usepackage[capitalize]{cleveref}
\crefname{section}{Sec.}{Secs.}
\Crefname{section}{Section}{Sections}
\Crefname{table}{Table}{Tables}
\crefname{table}{Tab.}{Tabs.}

\DeclareMathOperator*{\argmax}{argmax} 
\newcommand{\x}{{\bf x}}
\newcommand{\z}{{\bm z}}
\newcommand{\m}{{\bm m}}

\newcommand{\w}{{\bm w}}

\newcommand{\p}{{\bm p}}

\newcommand{\D}{\mathcal{D}}

\newcommand{\R}{\mathbb{R}}

\newcommand{\experimentsection}[1]{\subsection{#1}}
\newcommand{\bfname}[1]{{\bf #1}}
\newcommand{\name}{{\sc Fact }}
\newcommand{\mame}{{\sc Fact}}

\usepackage{accents}
\newlength{\dhatheight}
\newcommand{\doublehat}[1]{%
	\settoheight{\dhatheight}{\ensuremath{\hat{#1}}}%
	\addtolength{\dhatheight}{-0.35ex}%
	\hat{\vphantom{\rule{1pt}{\dhatheight}}%
		\smash{\hat{#1}}}}

\def\cvprPaperID{9011} 
\def\confName{CVPR}
\def\confYear{2022}

\begin{document}

\title{Forward Compatible Few-Shot Class-Incremental Learning}


\author{Da-Wei Zhou$^1$, Fu-Yun Wang$^1$, Han-Jia Ye$^1$\footnotemark[2], Liang Ma$^2$, Shiliang Pu$^2$, De-Chuan Zhan$^1$\\
	$^1$ State Key Laboratory for Novel Software Technology, Nanjing University
	$^2$ Hikvision Research Institute\\
{\small \{zhoudw, yehj, zhandc\}@lamda.nju.edu.cn, wangfuyun@smail.nju.edu.cn, \{maliang6, pushiliang.hri\}@hikvision.com}	
}

\maketitle

\footnotetext[2]{Correspondence to: Han-Jia Ye (yehj@lamda.nju.edu.cn)}

\begin{abstract}
	
	Novel classes frequently arise in our dynamically changing world, e.g., new users in the authentication system, and a machine learning model should recognize new classes without forgetting old ones. This scenario becomes more challenging when new class instances are insufficient, which is called few-shot class-incremental learning (FSCIL). 
	Current methods handle incremental learning retrospectively by making the updated model similar to the old one. By contrast, we suggest learning prospectively to prepare for future updates, and propose ForwArd Compatible Training (\mame) for FSCIL. 
	Forward compatibility requires future new classes to be easily incorporated into the current model based on the current stage data, and we seek to realize it by reserving embedding space for future new classes.
	In detail, we assign virtual prototypes to squeeze the embedding of known classes and reserve for new ones. Besides, we forecast possible new classes and prepare for the updating process.   The virtual prototypes allow the model to accept possible updates in the future, which act as proxies scattered among embedding space to build a stronger classifier during inference.
	\name efficiently incorporates new classes with forward compatibility and meanwhile resists forgetting of old ones.
	Extensive experiments validate  \mame's state-of-the-art performance. 
	Code is available at: \url{https://github.com/zhoudw-zdw/CVPR22-Fact}

\end{abstract}

\section{Introduction}

Recent years have witnessed the significant breakthroughs of deep neuron networks in many vision tasks~\cite{deng2009imagenet,tan2020efficientdet,he2015residual,simonyan2014very,ye2021contextualizing}. However, data often come in stream format~\cite{gomes2017survey} with emerging new classes~\cite{zhou2021learning,zhou2021learningtnnls,chou2020adaptive} in real-world applications, \eg, new types of products in e-commerce. It requires a model to incorporate new class knowledge incrementally, which is called Class-Incremental Learning (CIL). When updating the model with new classes, a fatal problem occurs, namely catastrophic forgetting~\cite{french1999catastrophic} --- the discriminability of old classes drastically declines. How to design effective CIL algorithms to overcome catastrophic forgetting has attracted much interest in the computer vision field~\cite{li2017learning,kirkpatrick2017overcoming,zhao2020maintaining,wu2019large,yan2021dynamically}.

\begin{figure}[t]
	\begin{center}
		\includegraphics[width=1\columnwidth]{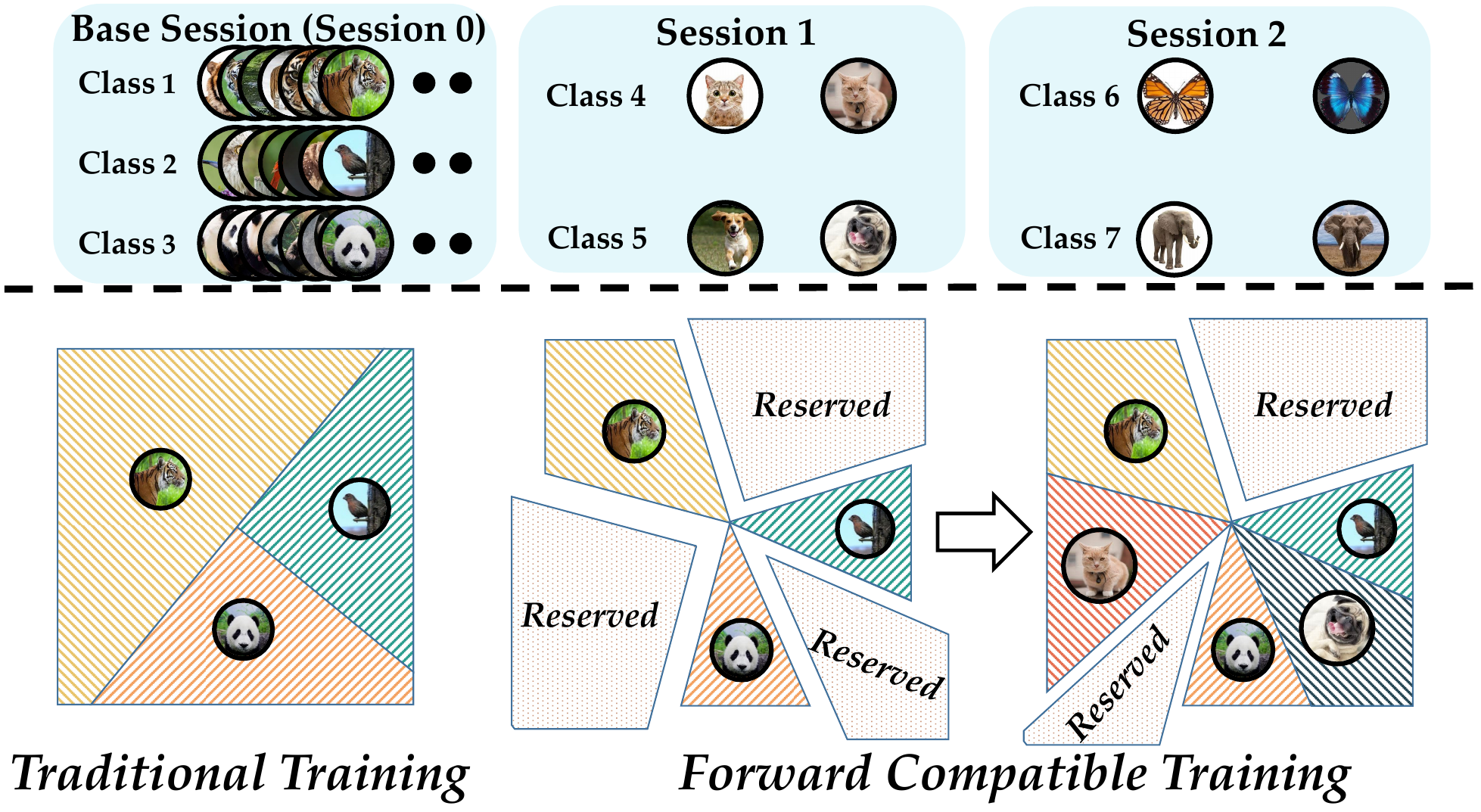}
	\end{center}
	\vspace{-6mm}
	\caption{\small   Top: the setting of FSCIL. 
		We need to maintain a classifier covering all classes, where sessions with non-overlapping classes arrive sequentially. Ample training instances are available in the base session, while only few-shot instances in incremental sessions.
		The model should incorporate new classes without forgetting old ones.  Bottom: forward compatible training scheme. Different from traditional training paradigm, we reserve the embedding space for new classes in the base session for future possible extensions. 	} \label{figure:intro}
	\vspace{-6mm}
\end{figure}

Current CIL methods address the scenario where new classes are available with \emph{sufficient} instances.  However, the data collection and labeling cost can be relatively high in many applications. Take a rare bird classification model for an example. We can only use few-shot images to train the incremental model since they are hard to collect. This task is called Few-Shot Class-Incremental Learning (FSCIL), which is shown at the top of Figure~\ref{figure:intro}. A model needs to sequentially incorporate new classes with \emph{limited} instances without harming the discriminability of old classes. Apart from the  forgetting problem, it also triggers overfitting on few-shot instances. As a result, some algorithms~\cite{zhang2021few,zhu2021self} are proposed to solve FSCIL from the few-shot learning perspective, aiming to alleviate overfitting in model updating.

In FSCIL, a sequence of models should work together with harmony, \ie, the updated model should maintain the discriminability of old classes. Such a learning process is similar to software development. The newer version software should accept the data that worked under the previous version, which is referred to as `backward compatibility'~\cite{varma2009backward,nagarakatte2009softbound}.
It measures the capability of different systems to work together without adaptation.
From this perspective, the ability to overcome forgetting represents model's backward compatibility --- if an updated model is good at classifying former classes, it is compatible with the old model and does not suffer forgetting. Consequently, CIL methods seek to increase backward compatibility by maintaining  discriminability of old classes, and FSCIL methods achieve this by fixing embedding module and incorporating new classes.

Current methods concentrate on backward compatibility, which shifts the burden of overcoming forgetting to the later model. However, if the former model works poorly, the latter model would degrade consequently. 
It is impossible to maintain backward compatibility with limited instances in incremental stages.
Take software development for an example. If an early version is poorly designed, the later version needs to work hard putting patches to maintain backward compatibility. By contrast, a better solution is to consider future extensions in the early version and reserve the interface in advance. Consequently, another compatibility, namely \emph{forward compatibility} is more proper for FSCIL, which prepares the model for possible future updates.

The model with forward compatibility should be \emph{growable and provident}. On the one hand, growable means the model  is aware of the incoming classes in the future and makes room for their embedding space. Hence, the model does not need to squeeze the space of former classes to make room for new ones when updating. Provident indicates the ability to forecast possible future classes. The model should anticipate the future and develop methods to minimize the effects of shocks and stresses of future events. Benefits from forward compatibility, the embedding space of old classes will be more compact, and new classes can be easily matched to the reserved space, as shown in Figure~\ref{figure:intro}.

In this paper, we propose ForwArd Compatible Training (\mame) for FSCIL to prepare the model for future classes. To make the model growable, we pre-assign multiple virtual prototypes in the embedding space, pretending they are the reserved space. By optimizing these virtual prototypes, we push the instances from the same class to be closer and reserve more spaces for incoming new classes.
Moreover, we also generate virtual instances via instance mixture to make the model provident. The virtual instances enable us to reserve the embedding space with explicit supervision. 
These reserved virtual prototypes can be seen as informative basis vectors during inference, with which we can build a strong classification model incrementally.
Vast experiments on benchmark datasets under various settings are conducted, validating the effectiveness of \mame.

\section{Related Work}

\noindent\bfname{Few-Shot Learning (FSL):} aims to fit unseen classes with insufficient training instances~\cite{wang2020generalizing,chen2018closer}.  Few-shot learning algorithms can be roughly divided into two groups: optimization-based methods and metric-based methods. Optimization-based methods try to enable fast model adaptation with few-shot data~\cite{finn2017model,arnold2021maml,antoniou2018train,nichol2018first}. Metric-based algorithms utilize a pretrained backbone for feature extraction, and employ proper distance metrics between support and query instances~\cite{snell2017prototypical,zhang2020deepemd,sung2018learning,vinyals2016matching,liu2020negative,ye2022identifying,ye2020heterogeneous}.

\noindent\bfname{Class-Incremental Learning (CIL):} aims to learn from a sequence of new classes without forgetting old ones, which is now widely discussed in various computer vision tasks~\cite{de2019continual,zhou2021co,zhou2021pycil,yang2021cost}. Current CIL algorithms can be roughly divided into three groups. The first group estimates the importance of each parameter and prevents important ones from being changed~\cite{kirkpatrick2017overcoming,aljundi2018memory,zenke2017continual}. The second group utilizes knowledge distillation to maintain the model's discriminability~\cite{hinton2015distilling,li2017learning,rebuffi2017icarl}. Other methods rehearsal former instances to overcome forgetting~\cite{wu2019large,zhao2020maintaining,belouadah2019il2m,zhu2021prototype}.
Pernici \etal \cite{pernici2021class} pre-allocates classifiers for future classes, which needs an extra memory for feature tuning and is unsuitable for FSCIL.

\noindent\bfname{Few-Shot Class-Incremental Learning:} is recently proposed to address the few-shot inputs in the incremental learning scenario~\cite{kukleva2021generalized,cheraghian2021synthesized,achituve2021gp,zhao2021mgsvf}. TOPIC~\cite{tao2020few} uses the neural gas structure to preserve the topology of features between old and new classes to resist forgetting.  Semantic-aware knowledge distillation~\cite{Cheraghian_2021_CVPR} treats the word embedding as auxiliary information, and builds knowledge distillation terms to resist forgetting. To resist overfitting on few-shot inputs, FSLL~\cite{mazumder2021few} selects a few parameters to be updated at every incremental session. CEC~\cite{zhang2021few} is the current state-of-the-art method, which utilizes an extra graph model to propagate context information between classifiers for adaptation.

\noindent\bfname{Backward Compatible Learning:} The concept `compatibility' is a design characteristic considered in software engineering~\cite{nagarakatte2009softbound,gheorghioiu2003interprocedural,varma2009backward,xu2004efficient}. Forward compatibility allows a system to accept input intended for a later version of itself, and backward compatibility allows for interoperability with an older legacy system.
They are introduced to the machine learning field in~\cite{bansal2019updates,srivastava2020empirical}. Recent work focuses on improving model's backward compatibility~\cite{shen2020towards,meng2021learning,budnik2021asymmetric}, while we are the first to address the model's forward compatibility in FSCIL.

\section{From Old Classes to New Classes}
In this section, we first describe the setting of FSCIL and then introduce the baseline methods and their limitations.

\subsection{Few-Shot Class-Incremental Learning}
\noindent\textbf{Base Session:} In FSCIL, a  model first receives the training set $\D^{0}=\left\{\left(\x_{i}, y_{i}\right)\right\}_{i=1}^{n_0}$ with \emph{sufficient} instances and evaluated with the testing set $\D^{0}_t=\left\{\left(\x_{j}, y_{j}\right)\right\}_{j=1}^{m_0}$.
$\D^0$ is called the base task.\footnote{We interchangeably use `task' and 'session' in this paper.}
$\x_i \in \R^D$ is a training instance of class $y_i \in Y_0$. $Y_0$ is the label space of base task. 
An algorithm fits a model $f(\x)$ to minimize the empirical risk over testing set:
\begin{align} \label{eq:closed} \textstyle
	\sum_{\left(\x_{j}, {y}_{j}\right) \in \D_t^{0}}
	\ell\left(f
	\left(\x_{j} 
	\right), {y}_{j}\right) \,,
\end{align}
where $\ell(\cdot,\cdot)$ measures the discrepancy between prediction and ground-truth label. 
The model can be decomposed into embedding and linear classifier: $f(\x)=W^{\top}\phi(\x)$, where $\phi(\cdot):\mathbb{R}^{D} \rightarrow \mathbb{R}^{d}$ and $W\in\mathbb{R}^{d\times |{Y}_{0}|}$.
We denote the classifier for class $k$ as $\w_k$: $W=[\w_1,\cdots,\w_{|Y_0|}]$.

\begin{figure*}[t]
	\vspace{-6mm}
	\begin{center}
		\includegraphics[width=1.7\columnwidth]{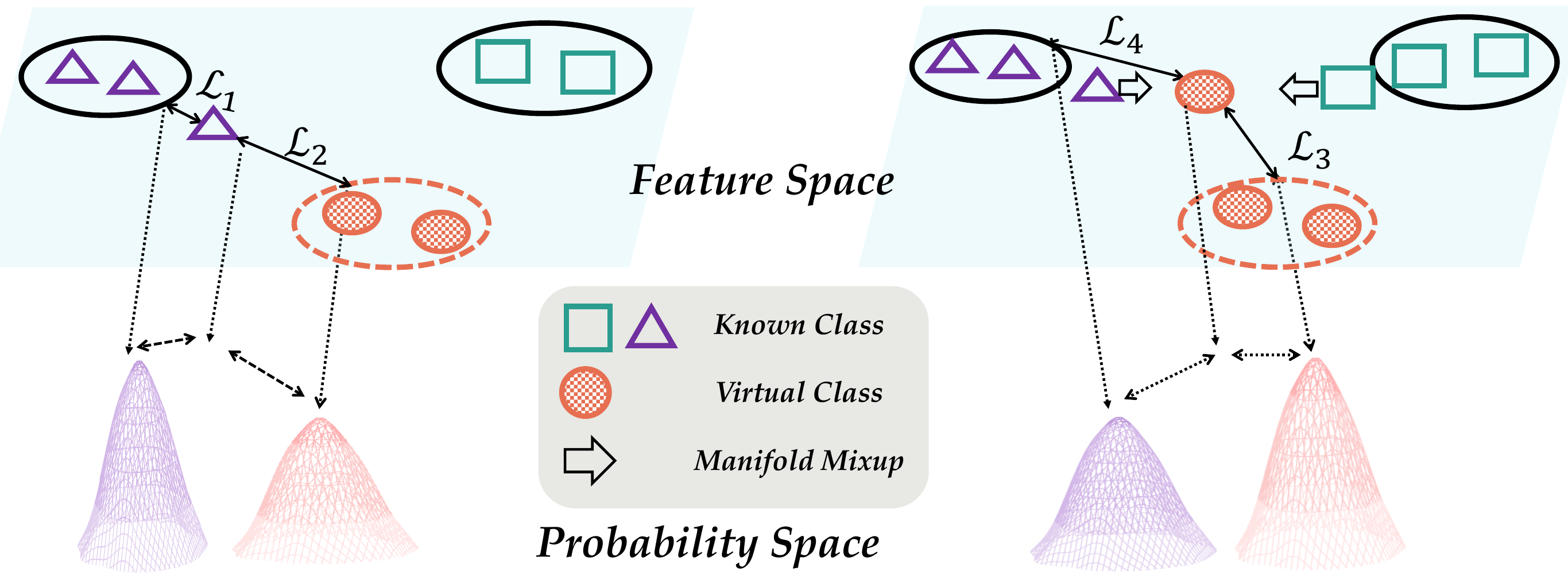}
	\end{center}
	\vspace{-6mm}
	\caption{\small  Illustration of \mame. Left: making the model growable. Apart from the cross-entropy loss ($\mathcal{L}_1$), the model also assigns an instance to a virtual class ($\mathcal{L}_2$), which reserves the space for new classes. Right: making the model provident. We first forecast virtual instances by manifold mixup (shown with arrow) and then conduct a symmetric reserving process by assigning it to the virtual class and known class. The training target is a bimodal distribution, which forces the instance to be assigned to different clusters and reserve embedding space.
	} \label{figure:teaser}
	\vspace{-5mm}
\end{figure*}

\noindent \textbf{Incremental Sessions:} New classes often arrive incrementally with \emph{insufficient} instances in practical applications, \ie, a sequence of  datasets $\left\{\D^{1}, \cdots, \D^{B}\right\}$ will emerge sequentially. $\D^{b}=\left\{\left(\x_{i}, y_{i}\right)\right\}_{i=1}^{NK}$, where $y_i \in Y_b$, and $Y_b$ is the label space of task $b$. 
$Y_b  \cap Y_{b^\prime} = \varnothing$ for $b\neq b^\prime$. 
We only have access to $\D^b$ when training task $b$.
The limited instances in each dataset can be organized as $N$-way $K$-shot format, \ie, there are $N$ classes in the dataset, and each class has $K$ examples.
Facing a new dataset $\D^b$, a model should learn new classes and maintain performance over old classes, \ie, minimize the empirical risk over all testing datasets:
\begin{align} \label{eq:fscil_risk}\textstyle
	\sum_{\left(\x_{j}, {y}_{j}\right) \in \D_t^{0}\cup \cdots \D_t^{b} }\ell \left(f
	\left(\x_{j}	\right), {y}_{j}\right) \,.
\end{align}

\subsection{Backward Compatible Training for FSCIL}
\label{sec:protonet}
\noindent\textbf{Knowledge Distillation:} To make the updated model still capable of classifying the old class instances, a common approach in CIL combines cross-entropy loss and knowledge distillation loss~\cite{hinton2015distilling}. It builds a mapping between the former and the current model to maintain backward compatibility:
\begin{align}\label{eq:icarl}	
	\begin{split}	 
		&\mathcal{L}(\x, y)=(1-\lambda) \mathcal{L}_\text{CE}(\x, y)+\lambda \mathcal{L}_\text{KD}(\x)\\ 
		&	\mathcal{L}_\text{KD}(\mathbf{x}) =  \sum_{k=1}^{|\mathcal{Y}_{b-1}|}-
		\mathcal{S}_k(\bar{W}^{\top}\bar{\phi}(\x))
		\log \mathcal{S}_k({W}^{\top}\phi(\x)) \,,
	\end{split}
\end{align}
where $\mathcal{Y}_{b-1}$ denotes the set of old classes, $\mathcal{Y}_{b-1}=Y_0\cup\cdots Y_{b-1}$, and $\mathcal{S}_k(\cdot)$ denotes the $k$-th class probability after softmax operation. $\bar{W}_{}$ and $\bar{\phi}$ correspond to frozen classifier and embedding before learning $\D^b$. Aligning the output of old model and current model helps to maintain model's discriminability and encourages backward compatibility.

\noindent\textbf{Prototypical Network:}
Knowledge distillation suffers from overfitting with few-shot inputs and works poorly. 
As a result, FSL algorithms are modified to conquer overfitting and encourage backward compatibility.
ProtoNet~\cite{snell2017prototypical}  trains the model on base classes with cross-entropy loss. The embedding $\phi(\cdot)$ is then fixed and utilized to extract average embedding (\ie, prototype) of each class:
\begin{align} \label{eq:prototype}\textstyle
	\p_i=\frac{1}{K}
	{\sum_{j=1}^{|\mathcal{{D}}^b|}\mathbb{I}(y_j=i)\phi(\x_j)}
	\,.
\end{align}
$\mathbb{I}(\cdot)$ is the indicator function. The averaged embedding represents the common pattern of each class, which is used as the classifier, \ie, $\w_i=\p_i$. The embedding is fixed throughout the incremental stages, and backward compatibility is maintained between old and new models.

\noindent\textbf{Ignorance of Forward Compatibility:}
Eq.~\ref{eq:icarl} and Eq.~\ref{eq:prototype} focus on different aspects of backward compatibility. The former maintains discriminability by aligning the old and new models into the same scale, and the latter fixes the embedding to prevent it from shifting away. 
They treat incremental learning from a \emph{retrospective} standpoint --- we need to keep the model unchanged during updating.
However, none of them consider the \emph{forward compatibility}, \ie, when training the model at the current session, it should prepare for the future possible updates.
It is profitable to treat incremental learning from a \emph{prospective} view --- look ahead to prepare for possible updates.
In other words, the training process should be `\emph{future-proof}'. If we reserve the embedding space for possible new classes and anticipate their possible pattern, the adaptation cost will be released in the future. 

\section{Forward Compatible Training for FSCIL}

Motivated by the potential of forward compatibility, we seek to enhance such characteristics in FSCIL. The training target is to look ahead and prepare for incoming classes in the base session, and we realize it from two aspects.
On the one hand, to make the model growable, we seek to optimize the posterior probability into bimodal distribution --- we assign each instance to an extra class except for the ground-truth class. 
The extra label stands for the reserved class space for new classes, which is optimized explicitly during training. 
On the other hand, to make the model provident, we try to foresee the possible distribution of new classes via instance mixture. The mimicked new instances help transform static training into incremental training. Forward compatibility is kept during this training paradigm.

We first introduce how to reserve embedding space, and then discuss how to use them for inference.

\subsection{Pretrain with Virtual Prototypes}

\subsubsection{Allocating Virtual Prototypes} 
In the base session, a common pretrain method is to optimize the empirical loss over the training set, which  takes no consideration of future incremental learning process and overspreads the embedding space. There are only $|Y_0|$ classes in  $\D^{0}$, but the model needs to handle $|Y_0|+NB$ classes in the final session. As a result, the traditional training paradigm needs to squeeze the embedding of old classes to make room for new ones, which takes many rounds and is unsuitable for FSCIL.
Our classification is based on the similarity between instance embedding and the class prototype (we replace the classifier weight with the class prototype), \ie, $p(y|\x) \propto \text{sim}\langle \w_y,\phi(\x)\rangle$, the more similar they are, the more probably $\x$ belongs to class $y$.
We use the cosine classifier as the similarity measure\footnote{We omit the norm for ease of discussion, \ie, $f(\mathbf{x})={W}^{\top}{\phi({\x})}$.} : $f(\mathbf{x})=(\frac{W}{\|W\|_{2}})^{\top}(\frac{\phi(\mathbf{x})}{\|\phi(\mathbf{x})\|_{2}})$. 
To enhance the forward compatibility, we pre-assign several virtual prototypes $P_v$ in the embedding space, and treat them as `virtual classes'. $P_v=[\p_1,\cdots,\p_V]\in\mathbb{R}^{d\times V}$, where $V$ is the number of virtual classes. Denote the output of current model as $f_v(\x)=[W,P_v]^\top\phi(\x)$, we seek to reserve the embedding space for these $V$ classes:
	\begin{align} \label{eq:dummy}
		\begin{split}
			\mathcal{L}_v(\x,y)=\underbrace{\ell (f_v(\x),y)}_{\mathcal{L}_1}	+\gamma\underbrace{\,\ell\left(\text{Mask}(f_v(\x),y),\hat{y}\right)}_{\mathcal{L}_2}\\
			\text{Mask}(f_v(\x),y)=f_v(\x) \otimes ({\bf 1}-\text{OneHot}(y)) \,,
		\end{split}
	\end{align}
	where $\hat{y}=\argmax_{v} \p_v^\top\phi(\x) + |Y_{0}|$ is the virtual class with maximum logit, acting as the pseudo label. $\otimes$ is Hadamard product (element-wise multiplication), ${\bf 1}$ is an all-ones vector. The first item in Eq.~\ref{eq:dummy} corresponds to the vanilla training loss, which matches the output to its ground-truth label. The second term first masks out the ground-truth logit with function $\text{Mask}(\cdot,\cdot)$, and then matches the rest part to the pseudo label $\hat{y}$. Since $\hat{y}$ is the virtual class with maximum logit, Eq.~\ref{eq:dummy} reserves the embedding space for $\hat{y}$ explicitly.
	
	\noindent\bfname{Effect of Virtual Prototypes:} 
	Eq.~\ref{eq:dummy} forces the output of $f_v(\x)$ to be bimodal, which is shown in left part of Figure~\ref{figure:teaser}. The first item forces an instance to be nearest to its ground-truth cluster, and the second term matches it to the nearest virtual cluster. 
	By optimizing Eq.~\ref{eq:dummy}, all non-target class prototypes will be pushed away from the reserved virtual prototype, and the decision boundary will be pushed in the same direction.
	As a result, the embedding of other classes will be more compact, and the embedding spaces for virtual classes will be reserved.
	Hence, the model becomes growable and forward compatibility is enhanced. We set V to the number of new classes, \ie, $V=NB$ as default.

	\subsubsection{Forecasting Virtual Instances}
	To make the model provident, we try to equip the model with `future-proof' ability --- if it has seen the novel patterns in the pretrain stage, the reserved space would be more suitable for incoming new classes. To this aim, we attempt to generate new classes by instance mixture and reserve the embedding space for these generated instances.
	
	Motivated by the intuition that interpolations between two different clusters are often regions of low-confidence predictions~\cite{verma2019manifold}, we seek to fuse two instances by manifold mixup~\cite{verma2019manifold} and treat the fused instance as a virtual new class. We decouple the embedding into two parts at the middle layer: $\phi(\x)=g(h(\x))$. For any instance couple from different classes in the mini-batch, say $(\x_i,\x_j), y_i\ne y_j$, we fuse the embedding of this pair as a virtual instance:
	\begin{align} \label{eq:mixup}
		\z = g\left[\lambda h(\x_i)+(1-\lambda)h(\x_j)\right]\,,
	\end{align}
	where $\lambda\in[0,1]$ is sampled from Beta distribution.
	Denote the current output for ${\z}$ is $f_v({\z})=[W,P_v]^\top{\z}$.
	Similar to the virtual loss in Eq.~\ref{eq:dummy}, we can build a \emph{symmetric} loss for the virtual instance $\z$ 
	to reserve embedding space:
	\begin{align} \label{eq:mixloss}
		\mathcal{L}_f(\z)=	\underbrace{\ell (f_v(\z),\hat{y})}_{\mathcal{L}_3}+\gamma\underbrace{\,\ell\left(\text{Mask}(f_v(\z),\hat{y}),\doublehat{y}\right)}_{\mathcal{L}_4} \,,
	\end{align}
	where $\hat{y}$ is the same as in Eq.~\ref{eq:dummy}, which is the pseudo label among virtual classes, and $\doublehat{y}=\argmax_{k} \w_k^\top\z $ is the pseudo label among current known classes. Note that the trade-off parameter $\gamma$ in Eq.~\ref{eq:mixloss} is the \emph{same} as in Eq.~\ref{eq:dummy}.
	
	\noindent\bfname{Effect of Virtual Instances:} 
	Eq.~\ref{eq:mixloss} is the symmetric form of Eq.~\ref{eq:dummy}. Generating virtual classes forecasts the possible distribution of incoming new classes, which is shown in the right part of Figure~\ref{figure:teaser}.
	The first item pushes the mixed instance $\z$ towards a  virtual prototype and away from other classes, reserving the space for the virtual class.  
	Besides, the second term pushes the mixed instance towards the nearest known class, which trades-off between known and virtual classes to prevent known classes from being over-squeezed.
	Hence, the model becomes provident by mimicking future instances, and forward compatibility is enhanced.

	\noindent\bfname{Why Virtual Loss Enhance Compatibility:} The final loss is combined of Eq.~\ref{eq:dummy} and Eq.~\ref{eq:mixloss}, \ie, $\mathcal{L}=\mathcal{L}_v+\mathcal{L}_f$, which has four loss terms. We analyze the gradient towards $\phi$ and $W$ to find out the intrinsic intuition hidden in virtual prototypes. Denote the probability of class $y$ after softmax operator as $a_y$, the first term in Eq.~\ref{eq:dummy} corresponds to $-\log a_y$, whose negative gradient w.r.t. $\phi(\x)$ is: $
	-\nabla_{\phi(\x)}{\mathcal L_1}=\bm w_y -\sum_{i=1}^{|  Y_0| + V}a_i\bm w_{i}$.\footnote{We treat $[W,P_v]$ as a uniform classifier and do not differentiate $\w$ and $\p$ in the analysis for ease of discussion.} It pushes the embedding towards the direction of ground-truth class center and away from other classes, including virtual classes. Similarly, we obtain the gradient $
	-\nabla_{\phi(\x)}{\mathcal L_2}=\w_{\hat{y}} -\sum_{i=1}^{|  Y_0| +V}a_i\bm w_{i}$. Note that the logit $a_y$ on ground-truth class is masked-out when optimizing ${\mathcal L_2}$, and thus we can push the embedding of ${\phi(\x)}$ towards its nearest virtual prototype without hurting the classification performance. These conclusions are consistent with the bimodal distribution in Figure~\ref{figure:teaser}. We conduct the same  analysis for $\mathcal{L}_3$ and $\mathcal{L}_4$. Assume $g(\x)$ as an identity function, \ie, $\phi(\x)=h(\x)$, the gradient yields: $
	-\nabla_{\phi({\x_i})}{\mathcal L_3}=\lambda \left( \bm w_{\hat{y}} -\sum_{k=1}^{| Y_0| +  V}a_k\w_{k} \right)
	$, $
	-\nabla_{\phi({\x_j})}{\mathcal L_3}=(1-\lambda) \left( \bm w_{\hat{y}} -\sum_{k=1}^{| Y_0| +  V}a_k\w_{k} \right)
	$.	It is similar to the effect of $\mathcal{L}_2$, which pushes all other non-target class prototypes away from the mixed components. As a result, the embeddings of known classes are optimized to be more compact, facilitating forward compatibility.

	\noindent\bfname{Pseudo Code:}
	We give the pseudo code of \name in the supplementary. In each mini-batch, we first calculate the virtual loss as Eq.~\ref{eq:dummy}. Afterward, we shuffle the dataset and conduct manifold mixup to calculate the forecasting loss in Eq.~\ref{eq:mixloss}. Note that we do not combine all possible pairs in the dataset (costs $\mathcal{O}(n*n)$), but instead, we only mix instances from different classes with the same index (costs $\mathcal{O}(n)$).

	\subsection{Incremental Inference with Virtual Prototypes}
	We have elaborated on the insight of virtual prototypes, and the left question is how to use them during inference. 
	The main idea is to treat them as bases encoded in the embedding space, and consider the possible influence of these bases on the prediction results.
	Each time an incremental dataset $\mathcal{D}^b$ arrives, we extract the prototype of these new classes as Eq.~\ref{eq:prototype}, and expand our classifiers: $W=[W;\w_i, i\in Y_b]$. We treat class prototypes as the representation of each class. Based on the law of total probability, we have:
	\begin{align} \label{eq:unfold}
		\begin{split}
			p\left(y_i| {\phi(\x)}\right)&=p(\bm w_{i}| {\phi(\x)})\\
			&=\sum_{\p_v \in  P_v}p(\bm w_i| \p_{v},{\phi(\x)})p(\p_{v}| {\phi(\x)}) \,,
		\end{split}
	\end{align}
	where $
	p(\p_{v}|{\phi(\x)})=\frac{\exp\left({\p_{v}}^\top {\phi(\x)}\right)}{\sum_{\p_{v} \in P_v}\exp\left({\p_{v}}^\top {\phi(\x)}\right)} 
	$. Eq.~\ref{eq:unfold} implies that we can consider the possible influence of all informative virtual prototypes to get the final prediction.
	For embedding $\phi(\x)$ from class $y_i$ with pseudo label $\hat{y}=v$, it shall follow a bimodal distribution between class $\bm w_i$ and $\p_{v}$, which is consistent with the training target. Hence, we assume $p\left(\phi(\mathbf{x})| \bm{w}_{i},\bm{p}_{v}\right)=\eta \mathcal N(\phi(\x)| \bm w_i,\Sigma)+(1-\eta)\mathcal N(\phi(\mathbf x)| \bm p_v,\Sigma)$. $\phi(\mathbf{x})$ follows a Gaussian mixture distribution, which is a linear superposition of  $\mathcal N(\phi(\x)| \bm w_i,\Sigma)$ and $\mathcal N(\phi(\x)| \bm p_v,\Sigma)$.
	According to Bayes' Theorem:
	\begin{align}
		p(\bm w_i|{\p_v,\phi(\x)} )=\frac{p({\phi(\x)}| \bm w_i, \p_v) p(\bm w_i| \p_v)}
		{\sum_{j=1}^{|\mathcal{Y}_b|}p({\phi(\x)}| \bm w_j, \p_v) p(\bm w_j| \p_v)} \,.
	\end{align}
	$|\mathcal{Y}_b|$ is the number of classes seen before. 
	The last term in denominator and numerator measures the similarity between class $\w_i$ and $\p_v$, and we assume it follows a Gaussian distribution, \ie, $ p(\bm w_i| \p_v)=\mathcal N(\bm w_i| \bm p_v, \Sigma)$. We have:
	{\small \begin{align} \label{eq:inference}
			&p(\bm w_i|  \p_v,{\phi(\x)})=\frac{p({\phi(\x)}| \bm w_i, \p_v) p(\bm w_i| \p_v)}
			{\sum_{j=1}^{|\mathcal{Y}_b|}p({\phi(\x)}| \bm w_j, \p_v) p(\bm w_j| \p_v)}\\ \notag
			&=\frac{ \text{Mix}_\eta\left(\m\left(\w_i,\phi\left(\mathbf x\right)\right),\bm m\left(\bm p_v,\phi\left(\mathbf x\right)\right)\right)
				\bm m(\bm w_i,\bm p_v)} {\sum_{j=1}^{\left|\mathcal Y_b\right|}
				\text{Mix}_\eta\left(\m\left(\w_j,\phi\left(\mathbf x\right)\right),\bm m\left(\bm p_v,\phi\left(\mathbf x\right)\right)\right)	 \bm m(\bm w_j,\bm p_v)} \,,
	\end{align}}where $\bm m\left(\bm \mu,\bm t\right)=\exp \left(\left(\Sigma^{-1} \bm \mu \right)^{\top} \bm t-\frac{1}{2} \bm \mu^{\top} \Sigma^{-1} \bm \mu\right)$, $\text{Mix}_\eta\left(a,b\right)=\eta  a+\left(1-\eta\right) b$. When $\phi(\x),\p,\w$ are normalized and $\Sigma=I$, Eq.~\ref{eq:inference} turns into:
	{\small \begin{align} 
					\frac{\text{Mix}_\eta\left(\exp \left(\bm {w}_{i}^{\top} \left(\phi\left(\mathbf{x}\right)+\bm p_v\right)\right),\exp \left(\bm {p}_{v}^{\top} \left(\phi\left(\mathbf{x}\right)+\bm w_i\right)\right)\right) }
					{\sum_{j=1}^{\left|\mathcal{Y}_{b}\right|}
						\text{Mix}_\eta\left(\exp \left(\bm {w}_{j}^{\top} \left(\phi\left(\mathbf{x}\right)+\bm p_v\right)\right),\exp \left(\bm {p}_{v}^{\top} \left(\phi\left(\mathbf{x}\right)+\bm w_j\right)\right)\right)}\,. 
			\end{align}}Specifically, when $\eta=1$, Eq.~\ref{eq:inference}  further degrades into: $\frac{\exp \left(\bm {w}_{i}^{\top} (\phi(\mathbf{x})+\bm p_v)\right)}{\sum_{j=1}^{\left|\mathcal{Y}_{b}\right|}\exp \left(\bm {w}_{j}^{\top} (\phi(\mathbf{x})+\bm p_v)\right) }$. It turns into ProtoNet (Eq.~\ref{eq:prototype}) when $ p(\bm w_i| \p_v)$ is a uniform distribution, which we analyze in the supplementary.
			In our implementation, we set $\eta=0.5$ as default for simplification. The inference process builds a channel for virtual prototypes to be taken into account for the final prediction and reflects their influences.
			
			To summarize, with the help of virtual prototypes learned in the pretraining stage, we are able to obtain a more informative distribution for $\phi(\x)$, which helps build a stronger incremental  classifier in the incremental stage. The inference process validates the forward compatibility of \mame.

\begin{table*}[t] 
	\vspace{-5mm}
	\centering{
		\caption{\small Detailed accuracy of each incremental session on CUB200 dataset.
			The results of compared methods are cited from~\cite{tao2020few} and~\cite{zhang2021few}. Methods with $\dagger$ are reproduced under FSCIL setting with the source code. Please refer to supplementary for results on other datasets.
		}
		\vspace{-3.5mm}
		\resizebox{\textwidth}{!}{
			
			\begin{tabular}{llllllllllllll}
				\toprule
				\multicolumn{1}{c}{\multirow{2}{*}{Method}} & \multicolumn{11}{c}{Accuracy in each session (\%) $\uparrow$} & \multirow{2}{*}{PD $\downarrow$} & \multirow{2}{*}{$\Delta$ PD }\\ \cmidrule{2-12}
				\multicolumn{1}{c}{} & 0   & 1      & 2      & 3    & 4     & 5  & 6     & 7      & 8     &   9  & 10   &  &   \\ \midrule
				Finetune                & 68.68   & 43.70      & 25.05    & 17.72   & 18.08     & 16.95  & 15.10  & 10.06     & 8.93   &8.93  & 8.47   & 60.21&\bf +41.25    \\
				Pre-Allocated RPC$^\dagger$~\cite{pernici2021class}          & 68.47   & 51.00     &45.42    & 40.76  &   35.90   & 33.18  & 27.23  & 24.24     & 21.18   &17.34 & 16.20  & 52.27&\bf +33.31    \\
				iCaRL~\cite{rebuffi2017icarl}        & 68.68   & 52.65     & 48.61    & 44.16  & 36.62   & 29.52  & 27.83  & 26.26    & 24.01   &23.89  & 21.16  & 47.52&\bf +28.56   \\
				EEIL~\cite{castro2018end}         & 68.68   & 53.63      &47.91    & 44.20  & 36.30     & 27.46  & 25.93  & 24.70     & 23.95   &24.13 & 22.11  & 46.57&\bf +27.61    \\
				
				Rebalancing~\cite{hou2019learning}         & 68.68   & 57.12     & 44.21    & 28.78  & 26.71    & 25.66  & 24.62  & 21.52    & 20.12   &20.06  & 19.87  & 48.81&\bf +29.85    \\
				TOPIC~\cite{tao2020few}            & 68.68   & 62.49      & 54.81    & 49.99   & 45.25     & 41.40 & 38.35  & 35.36    & 32.22  &28.31 & 26.26   & 42.40&\bf +23.44      \\
				SPPR~\cite{zhu2021self}  & 68.68   & 61.85     & 57.43    & 52.68   & 50.19   & 46.88 & 44.65   & 43.07      & 40.17     & 39.63   & 37.33& 31.35 & \bf +12.39      \\
				Decoupled-NegCosine$^\dagger$~\cite{liu2020negative}    &74.96  & 70.57     & 66.62   & 61.32   & 60.09    & 56.06  & 55.03  & 52.78     & 51.50   &50.08 & 48.47   & 26.49&\bf +7.53     \\
				Decoupled-Cosine~\cite{vinyals2016matching}    &75.52  & 70.95     & 66.46   & 61.20   & 60.86    & 56.88  & 55.40  & 53.49     & 51.94   &50.93 & 49.31   & 26.21&\bf +7.25     \\
				Decoupled-DeepEMD~\cite{zhang2020deepemd}     & 75.35   & 70.69     & 66.68   & 62.34  & 59.76     & 56.54  & 54.61  & 52.52    &50.73   &49.20 & 47.60   & 27.75&\bf +8.79    \\
				CEC~\cite{zhang2021few}                & 75.85   & 71.94    & 68.50   & 63.50   & 62.43    & 58.27 & 57.73 & 55.81    &54.83  &53.52  & 52.28   & 23.57&\bf +4.61   \\
				\midrule
				\name             & \bf 75.90   & \bf 73.23     & \bf70.84    & \bf66.13  & \bf65.56   &\bf62.15  & \bf61.74   &\bf 59.83     & \bf58.41   & \bf57.89   & \bf56.94&\bf 18.96 \\

				\bottomrule
			\end{tabular}
		}\label{table:cub}
		\vspace{-4mm}
	}
\end{table*}

\begin{figure*}[t]
	\begin{center}
		\begin{subfigure}{0.685\columnwidth}
			\includegraphics[width=\columnwidth]{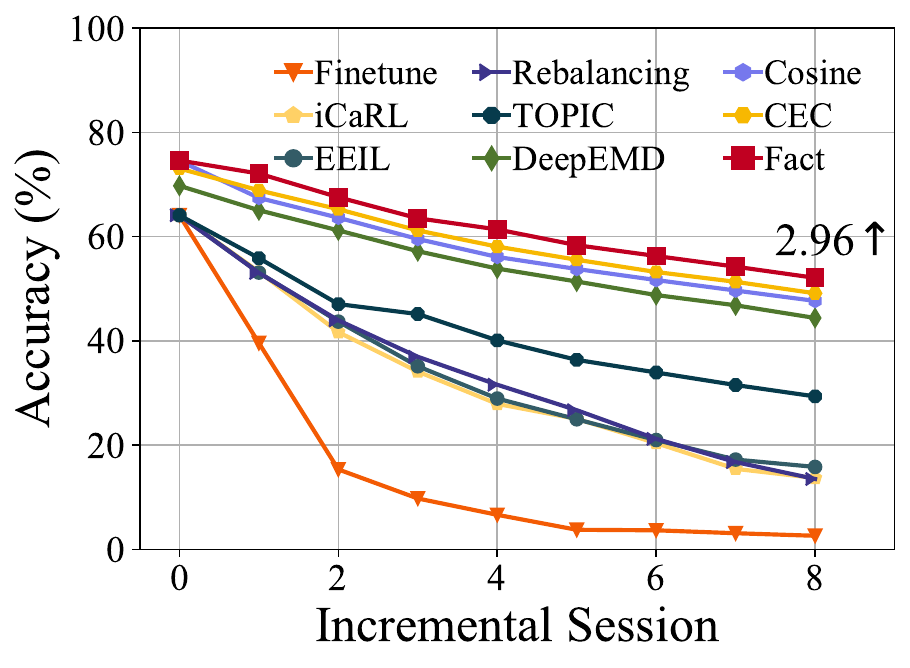}
			\caption{CIFAR100}	\label{figure:bmcifar}
		\end{subfigure}
		\begin{subfigure}{0.685\columnwidth}
			\includegraphics[width=\columnwidth]{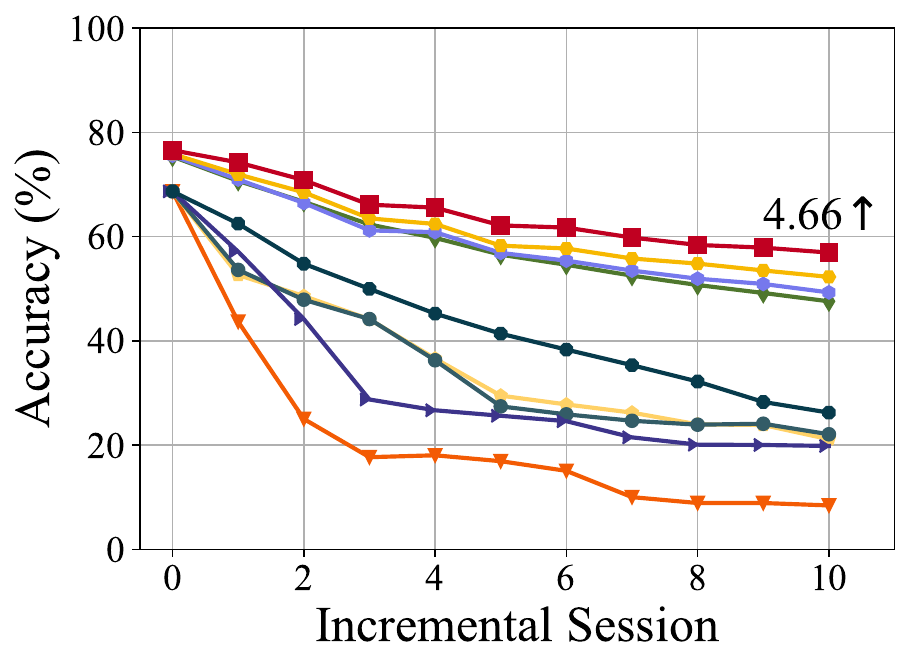}
			\caption{CUB200}	\label{figure:bmcub}
		\end{subfigure}
		\begin{subfigure}{0.685\columnwidth}
			\includegraphics[width=\columnwidth]{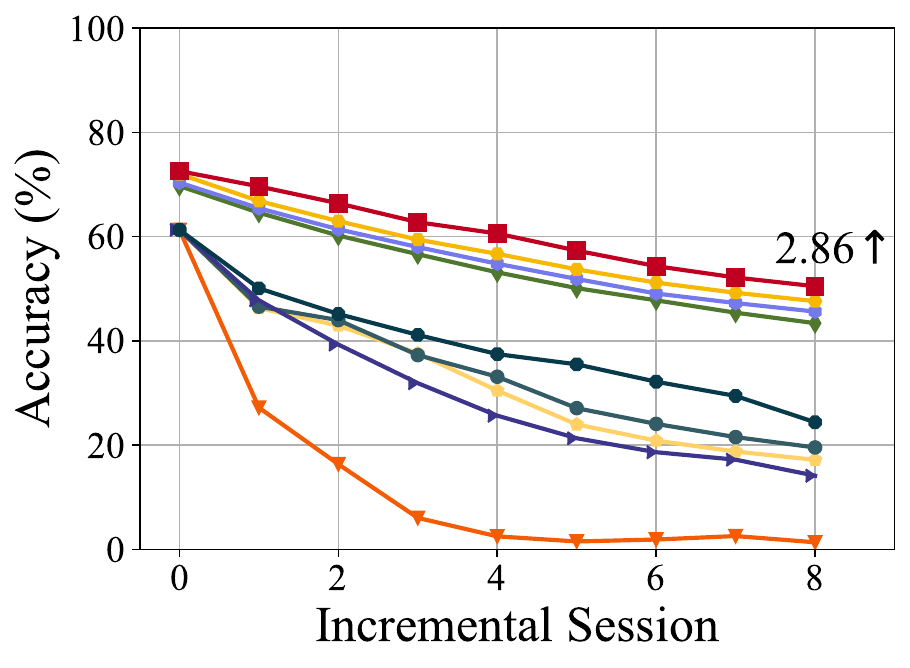}
			\caption{\textit{mini}ImageNet}	\label{figure:bmmini}
		\end{subfigure}
	\end{center}
	\vspace{-7mm}
	\caption{\small  Top-1 accuracy of each incremental session. We show the legends in (a), and annotate the performance gap after the last session between \name and the runner-up method at the end of each curve. Please refer to Table~\ref{table:cub} and supplementary for detailed values.
		\vspace{-7mm}
	} \label{figure:benchmark}
	
\end{figure*}

\section{Experiment}
In this section, we compare \name on benchmark FSCIL datasets with state-of-the-art methods and large scale dataset ImageNet. The ablations verify the effectiveness of forward compatible training, and we visualize the incremental process of \name with new classes.

\subsection{Implementation Details}
\noindent {\bf Dataset}: Following the benchmark setting ~\cite{tao2020few}, we evaluate the performance on CIFAR100~\cite{krizhevsky2009learning}, CUB200-2011~\cite{WahCUB2002011} and {\it mini}ImageNet~\cite{russakovsky2015imagenet}. We also experiment on the large scale dataset ImageNet ILSVRC2012~\cite{deng2009imagenet}. 
CIFAR100 contains 60,000 images from 100 classes.	 {CUB200} is a fine-grained image classification task with 200 classes. 
\textit {mini}ImageNet  is a subset of ImageNet~\cite{deng2009imagenet} with 100 classes. ImageNet1000  contains 1,000 classes, and we also sample a subset of 100 classes according to~\cite{wu2019large}, denoted as ImageNet100. 

\noindent {\bf Dataset Split:} For CIFAR100, \textit {mini}ImageNet and ImageNet100, 100 classes are divided into 60 base classes and 40 new classes. The new classes are formulated into eight \emph{5-way 5-shot} incremental tasks. For CUB200, 200 classes are divided into 100 base classes and 100 incremental classes, and the new classes are formulated into ten 10-way 5-shot incremental tasks. For ImageNet1000, 600 classes are selected as base classes, and the other 400 classes are formulated into eight 50-way 5-shot tasks.
We use the \emph{identical} training splits~\cite{tao2020few} (including base and incremental sessions) for every compared method for a fair comparison. The testing set is the same as the original one to evaluate holistically.

\begin{figure}[t]
	\begin{center}
		\begin{subfigure}{0.495\linewidth}
			\includegraphics[width=\columnwidth]{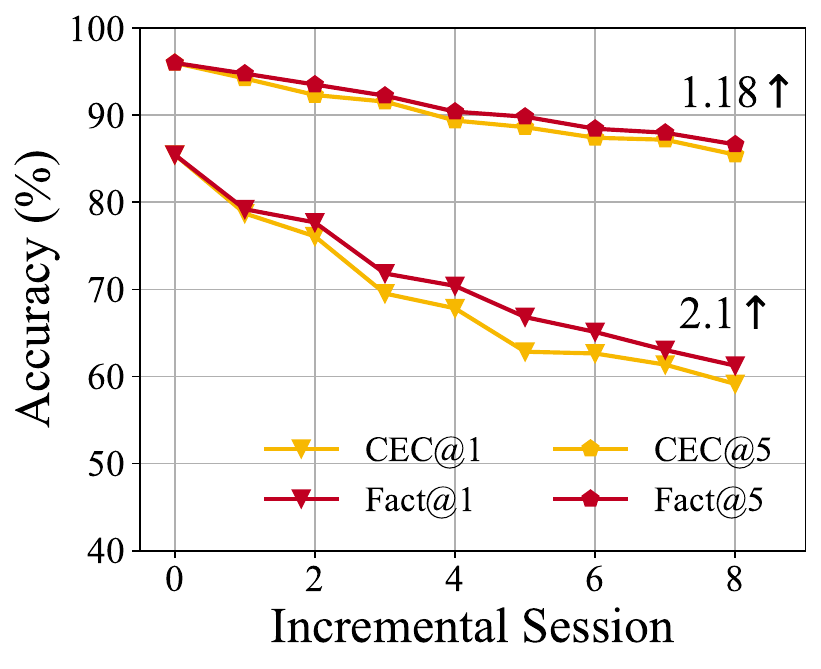}\label{figure:bmimage100}
			\caption{ImageNet100}	
		\end{subfigure}
		\hfill
		\begin{subfigure}{0.495\linewidth}
			\includegraphics[width=\columnwidth]{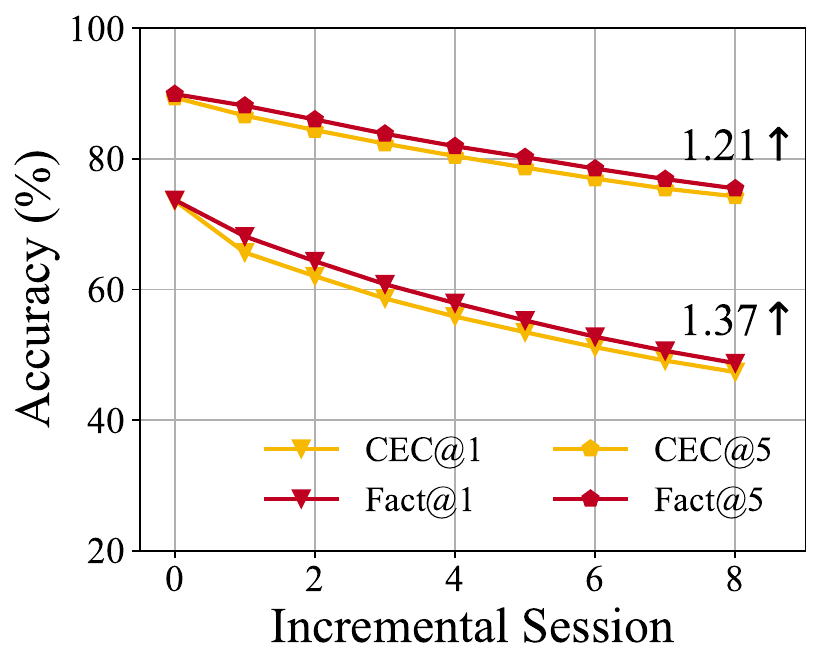}\label{figure:bmimage1000}
			\caption{ImageNet1000}	
		\end{subfigure}
		
	\end{center}
	\vspace{-7mm}
	\caption{\small  Top-1 and Top-5 accuracy on ImageNet100/1000. The performance gap is annotated at the end of each curve.
	} 
	\vspace{-5mm}\label{figure:imagenet}
\end{figure}

\noindent {\bf Compared methods:} We first compare to classical CIL methods iCaRL~\cite{rebuffi2017icarl}, EEIL~\cite{castro2018end} Pre-Allocated RPC~\cite{pernici2021class} and Rebalancing~\cite{hou2019learning}. Besides, we also compare to current SOTA FSCIL algorithms: TOPIC~\cite{tao2020few}, SPPR~\cite{zhu2021self}, Decoupled-DeepEMD/Cosine/NegCosine~\cite{zhang2020deepemd,vinyals2016matching,liu2020negative} and CEC~\cite{zhang2021few}. We report the baseline method by finetuning the model with few-shot instances as `finetune'.

\noindent {\bf Training details:} 
All models are deployed with PyTorch~\cite{paszke2019pytorch}.
We use the \emph{same} network backbone~\cite{tao2020few} for \emph{all} compared methods. For CIFAR100, we use ResNet20~\cite{he2015residual}, while for the others we use ResNet18.
The model is trained with a batch size of 256 for 600 epochs, and we use SGD with momentum for optimization. 
The learning rate starts from $0.1$ and decays with
cosine annealing.

\noindent {\bf Evaluation protocol:}  Following~\cite{tao2020few}, we denote the Top-1 accuracy after the i-th session as $\mathcal{A}_i$. We also quantitatively measure the forgetting phenomena with performance dropping rate (PD), \ie, $\text{PD}=\mathcal{A}_0-\mathcal{A}_B$, where $\mathcal{A}_0$ stands for the accuracy after base session, and $\mathcal{A}_B$ is the accuracy after the last session.

\begin{figure}[t]
	\vspace{-5mm}
	\begin{center}
		\includegraphics[width=0.9\columnwidth]{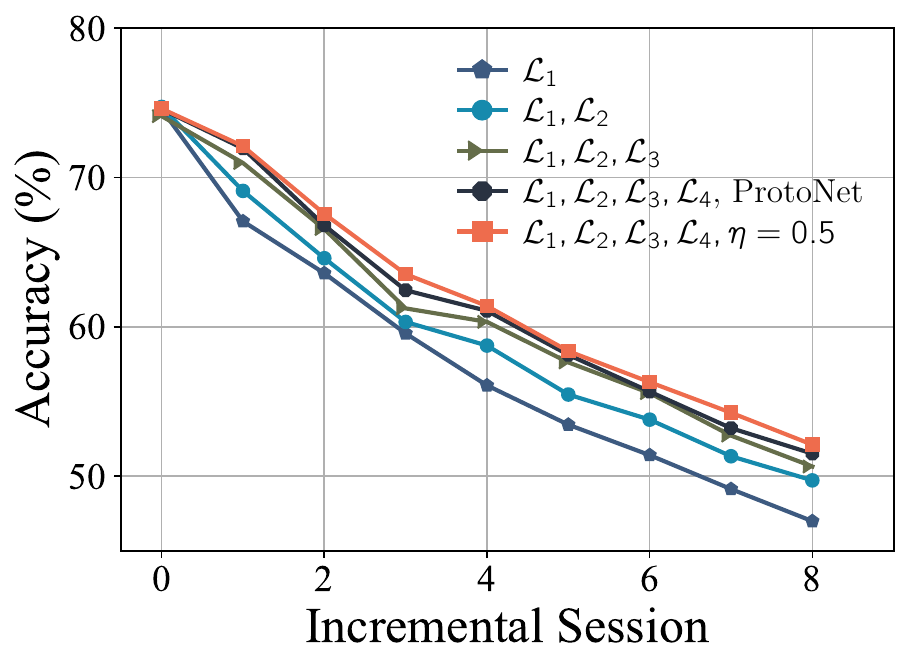}
	\end{center}
	\vspace{-7mm}
	\caption{\small  Ablation study. Reserving and forecasting the embedding space for new classes help enhance the forward compatibility.
		Every part in \name improves the performance of FSCIL.
	}
	\vspace{-5mm} \label{figure:ablation}
\end{figure}

\experimentsection{Benchmark Comparison}
We report the performance curve over benchmark datasets~\cite{tao2020few} \ie, CIFAR100, CUB200, and \textit{mini}ImageNet in Figure~\ref{figure:benchmark} and Table~\ref{table:cub}. We also report the results on large scale dataset, \ie, ImageNet100/1000 in Figure~\ref{figure:imagenet}.

We can infer from Figure~\ref{figure:benchmark} that \name consistently outperforms the current SOTA method, \ie, CEC, by 3-5\% on benchmark datasets.
The poor performance of CIL methods indicates that addressing the backward compatibility via output restriction is improper for FSCIL. The backward compatible tuning consumes numerous training instances, which is inconsistent with FSCIL scenario. \name also outperforms FSCIL methods which enhance backward compatibility by embedding freezing, indicating that considering forward compatibility is more proper and suitable for FSCIL. Besides, \name has better performance than Decoupled-NegCosine, which considers forward compatibility from another perspective. It reveals that our training scheme is more proper for FSCIL.
We report the detailed value on CUB200 dataset in Table~\ref{table:cub}. It shows that \name has a minimum degradation in terms of PD metric, indicating forward compatible training consistently resists forgetting in FSCIL.

Apart from the benchmark setting with small scale datasets, we also suggest conducting experiments with large scale datasets, \ie, ImageNet. We report top-1 and top-5 accuracy of \name and the runner-up method CEC in Figure~\ref{figure:imagenet}. As can be seen from the figures, \name still surpasses CEC by all metrics. To conclude, \name consistently handles small scale and large scale FSCIL tasks with SOTA performance.

\begin{figure}[t]
	\vspace{-5mm}
	\begin{center}
		\begin{subfigure}{0.48\columnwidth} 
			\includegraphics[width=\columnwidth]{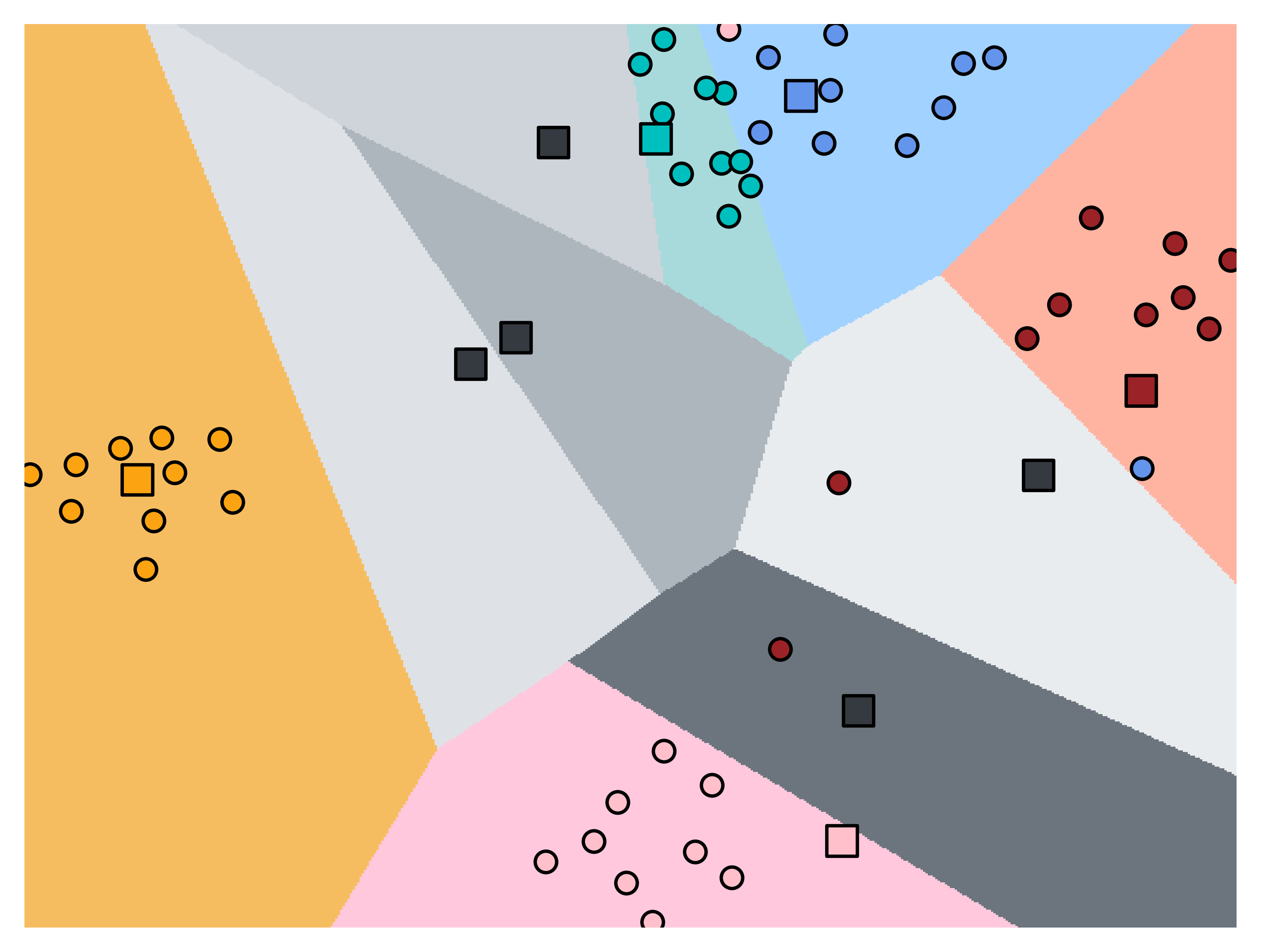} 
			\caption{Base session, 5 old classes \& 5 virtual prototypes.} \label{figure:tsne1}	
		\end{subfigure}
		\hfill
		\begin{subfigure}{0.48\columnwidth} 
			\includegraphics[width=\columnwidth]{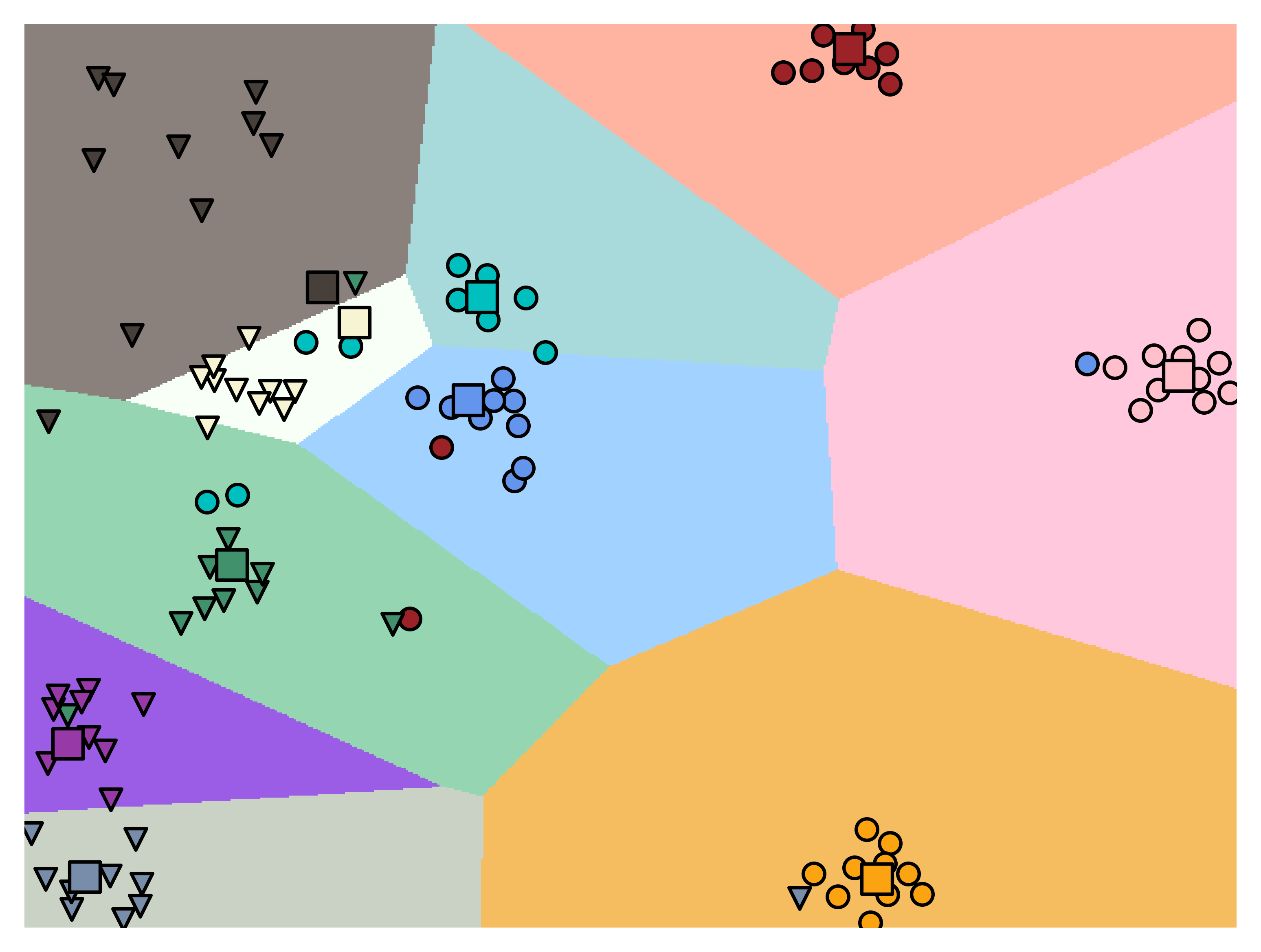} 
			\caption{Incremental session, 5 old classes \& 5 new classes.} \label{figure:tsne2}	
		\end{subfigure}
		\begin{subfigure}{0.8\columnwidth}
			\includegraphics[width=\columnwidth]{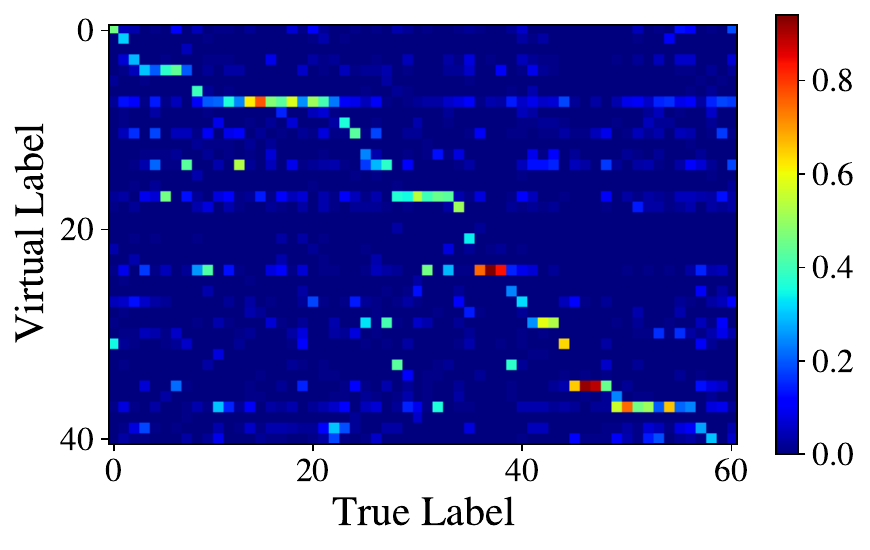}
			\caption{Pseudo label $\hat{y}$ assignment of each known class.}	 \label{figure:assign}
		\end{subfigure}
		
	\end{center}
	\vspace{-7mm}
	\caption{\small Top: visualization of the  decision boundary on CIFAR100 between 2 sessions. Old classes are shown in dots, and new classes are shown in triangles. The shadow region represents the decision boundary. Bottom: the pseudo label assignment for each known class. Each known class is assigned to one or more virtual classes.
	} \label{figure:tsne}
	\vspace{-5mm}
\end{figure}

\experimentsection{Ablation Study}
There are four loss terms in \mame, and we conduct an ablation study to analyze the importance of each component in \name on CIFAR100 dataset. The implementation details are the same as in the benchmark setting. We report the accuracy curve of different variants in Figure~\ref{figure:ablation}.

The model trained with $\mathcal{L}_1$  gets the lowest accuracy among all variants. $\mathcal{L}_1$ does not consider the future possible new classes and overspreads the embedding space with known classes. As a result, the model trained with $\mathcal{L}_1$ lacks forward compatibility and works poorly. When equipped with $\mathcal{L}_2$, the model pre-assigns extra virtual prototypes, which reserves the embedding space for new classes by virtual loss. Consequently, the model becomes growable and increases the forward compatibility. We then add $\mathcal{L}_3$ to forecast the possible new classes during the pretraining process. By anticipating the future and reserving the corresponding space, the model becomes provident with better compatibility. Optimizing $\mathcal{L}_4$ makes the final loss symmetric, which also increases the compactness of embedding and facilitates FSCIL. Lastly, these virtual prototypes can be utilized as bases encoded in the embedding space, which helps inference via Eq.~\ref{eq:inference}. The comparison between ProtoNet and Eq.~\ref{eq:inference} with $\eta=0.5$ verifies the effectiveness of these virtual prototypes during inference.
Ablations validate that forward compatible training is helpful for FSCIL.

\experimentsection{Visualization of Incremental Sessions}
We visualize the learned decision boundaries with t-SNE~\cite{van2008visualizing} on CIFAR100 dataset in Figure~\ref{figure:tsne1},\ref{figure:tsne2}. 
Figure~\ref{figure:tsne1} stands for the decision boundary in the base session, where we train 5 old classes and 5 virtual prototypes. Known classes are shown in colorful dots, and their prototypes are represented by squares.
Black squares represent virtual prototypes,
and their embedding space is shown in gray with different shades. As we can infer from Figure~\ref{figure:tsne1}, forward compatible training reserves the embedding space for new classes (the gray parts) and makes the known class embedding more compact. When moving to Figure~\ref{figure:tsne2}, we show the incremental stage when 5 new classes emerge (represented by triangles). The reserved space benefits the incremental learning process, and new classes do not need to squeeze the embedding of old ones during incremental learning process.

We record the pseudo label assigned for each class during training and show the frequency matrix in Figure~\ref{figure:assign}. The matrix is \emph{reconstructed} by switching lines to make similar classes close. As we can infer from the figure, each known class is assigned to one or more virtual classes.  Virtual prototypes cover all the known classes and prepare for the possible update between them.

\begin{figure}[t]
	\vspace{-4mm}
	\begin{center}
		\begin{subfigure}{0.47\linewidth}
			\includegraphics[width=\columnwidth]{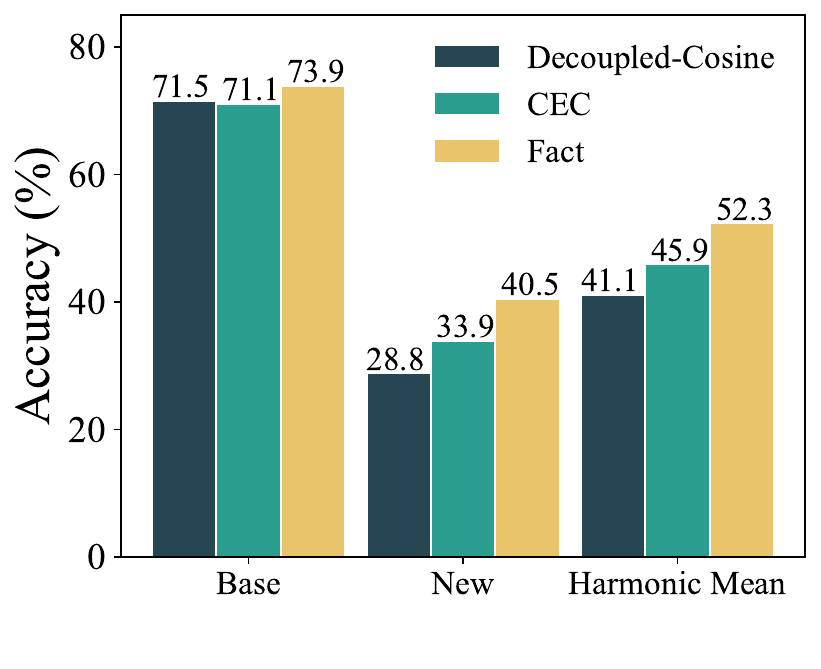}
			\caption{Accuracy of old and new classes}	\label{figure:ablaa}
		\end{subfigure}
		\hfill
		\begin{subfigure}{0.47\linewidth}
			\includegraphics[width=\columnwidth]{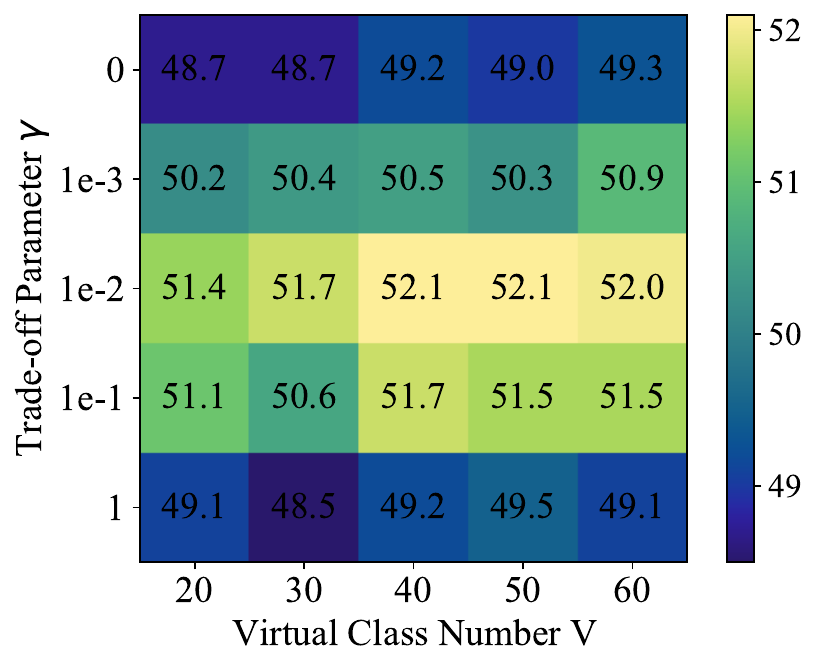}
			\caption{Influence of hyper-parameter}\label{figure:ablab}
		\end{subfigure}
		\vspace{-3mm}
		\begin{subfigure}{0.47\linewidth}
			\includegraphics[width=\columnwidth]{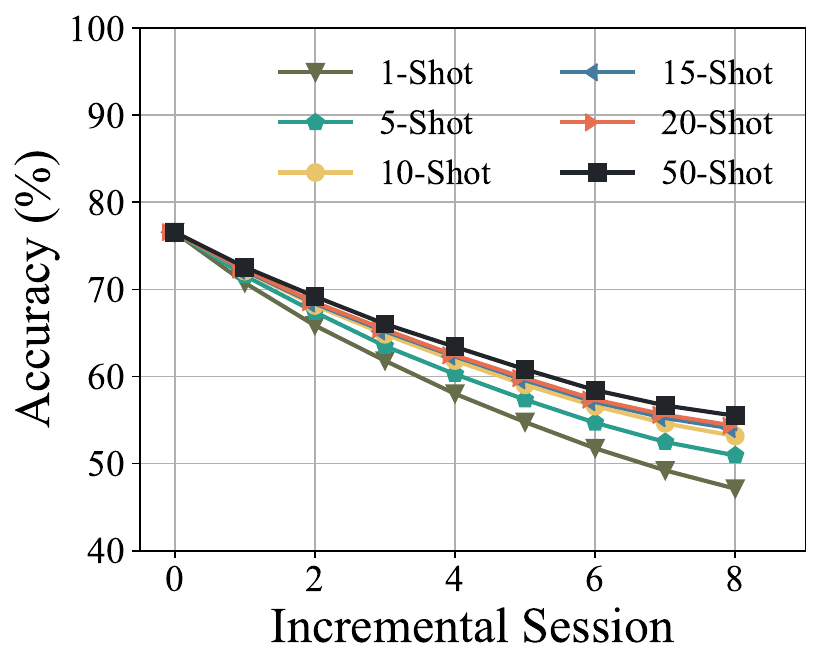}
			\caption{Influence of incremental shot}	\label{figure:ablac}
		\end{subfigure}
		\hfill
		\begin{subfigure}{0.47\linewidth}
			\includegraphics[width=\columnwidth]{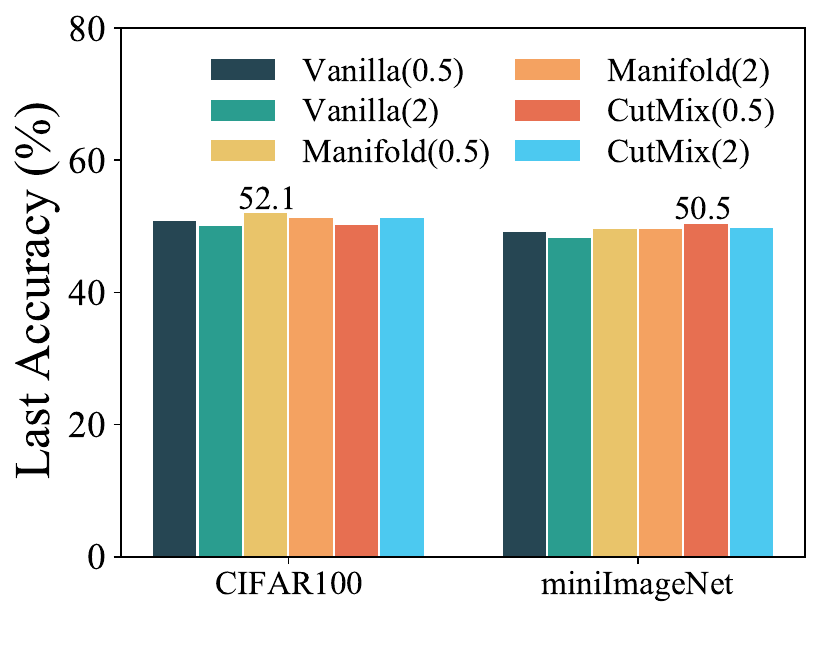}
			\caption{Influence of instance mixture}	\label{figure:ablad}
		\end{subfigure}
	\end{center}
	\vspace{-4mm}
	\caption{\small  Analysis on different performance measure, influence of hyper-parameters, incremental shot and instance mixture method.
	}\vspace{-6mm} \label{figure:analysis}
\end{figure}

\experimentsection{Further Analysis} \label{sec:hyper}

\noindent\textbf{Performance Measure:} The accuracy measure covers both old and new classes. To explore the ability of new class learning and forgetting resistance,
we report the accuracy of base classes and new classes, and the harmonic mean between them on CUB200. The other settings are the same as the benchmark, and the results are shown in Figure~\ref{figure:ablaa}. We also report the results of two competitive methods, \ie, Decoupled-Cosine and CEC. We can infer that \name has better performance on new classes, which verifies the effectiveness of forward compatible training.

\noindent\textbf{Hyper-Parameter:} There are two hyper-parameters in \mame, \ie, trade-off parameter $\gamma$ in Eq.~\ref{eq:dummy} and the number of virtual prototypes $V$.
We report the final accuracy on CIFAR100 varying the $\gamma$ and $V$ in Figure~\ref{figure:ablab}. We fix the other settings the same as in the benchmark experiment, and change $\gamma$ from $\{0,1e-3,1e-2,1e-1,1\}$. We also change the virtual class number from $\{20,30,40,50,60\}$, resulting in 25 compared results. We can infer from Figure~\ref{figure:ablab} that \name prefers small $\gamma$ and large $V$. However, since the increasing trend w.r.t. $V$ is trivial for $V>40$, we suggest to set $\gamma=0.01$ and $V=NB$ as default, where $NB$ indicates the number of new classes in total.

\noindent\textbf{Incremental Shot:} \name relies on the $N$-way $K$-shot dataset to estimate the class center of new classes, and we change the shot number to find out the influence on the last accuracy. We fix the incremental way the same as in the benchmark setting, and change the shot number $K$ among $\{1,5,10,15,20,50\}$ on \textit{mini}ImageNet. As we can infer from Figure~\ref{figure:ablac}, with more instances per class, the estimation of prototypes will be more precise, and the performance will correspondingly improve. However, we also find that the improvement trend becomes trivial for $K>20$.

\noindent\textbf{Instance Mixture Method:} Apart from manifold mixup~\cite{verma2019manifold}, there are other ways for instance mixture, \eg, vanilla mixup~\cite{zhang2018mixup} and CutMix~\cite{yun2019cutmix}. We explore the influence of the last accuracy by changing these mixture methods in Figure~\ref{figure:ablad}. The accuracy is annotated for the method with the best performance.
Since the mixture weight $\lambda$ is sampled from $\text{Beta}(\alpha,\alpha)$, we also change $\alpha$ from $\{0.5,2\}$. Figure~\ref{figure:ablad} implies that the final accuracy is robust to mixture methods, and we use manifold mixup with  $\alpha=0.5$ as default.

\section{Conclusion}

Few-shot class-incremental learning ability becomes an essential characteristic in real-world learning systems, which requires a model to learn few-shot new classes without forgetting old ones. 
In this paper, we address the importance of building forward compatible models for FSCIL. We pre-assign virtual prototypes in the embedding space to reserve the space for new classes from two aspects. The model becomes growable and provident by reserving and forecasting the future possible new classes, which releases the shock during model updating. These virtual prototypes act as bases in the embedding space and facilitate inference performance of FSCIL.
\name efficiently incorporates new knowledge into the current model and obtains SOTA performance.

\noindent\textbf{Limitations:} Possible limitations include the hypothesis of new class instances. It turns into CIL problem if sufficient new class instances are available, where forward and backward compatibility should be considered simultaneously.

\section*{Acknowledgments}

This research was supported by National Key
R\&D Program of China (2020AAA0109401), NSFC (61773198, 61921006,62006112), NSFC-NRF Joint Research Project under Grant 61861146001, Nanjing University Innovation Program for
Ph.D. candidate (CXYJ21-53), Collaborative Innovation Center of Novel Software Technology and Industrialization, NSF of Jiangsu Province (BK20200313), CCF-Hikvision Open Fund (20210005).

{\small
	\bibliographystyle{ieee_fullname}
	\bibliography{cvpr}
}

\setcounter{section}{0}
\renewcommand{\thesection}{\Roman{section}}
\begin{center}
	\textbf{\large Supplementary Material }
\end{center}
\setcounter{equation}{0}
\setcounter{figure}{0}
\setcounter{table}{0}
\setcounter{page}{1}
\makeatletter

\newcommand{\hathat}[1]{%
	\settoheight{\dhatheight}{\ensuremath{\hat{#1}}}%
	\addtolength{\dhatheight}{-0.65ex}%
	\hat{\vphantom{\rule{1pt}{\dhatheight}}%
		\smash{\hat{#1}}}}
\newcommand{\hathathat}[1]{%
	\settoheight{\dhatheight}{\ensuremath{\hat{#1}}}%
	\addtolength{\dhatheight}{-0.9ex}%
	\hat{\vphantom{\rule{1pt}{\dhatheight}}%
		\smash{\hat{#1}}}}

\section{Gradient Analysis}\label{sec:gradient}
In the main paper, we give the analysis about $\mathcal{L}_1$, $\mathcal{L}_2$, $\mathcal{L}_3$, $\mathcal{L}_4$ with regard to the embedding $\phi(\x)$. In this section, we give the full analysis about gradients, including the gradients with regard to $W$, and the scenario where $g(\cdot)$ is not an identity function.

With a bit of redundancy, we first revisit the notations defined in the main paper. We define the model output as $f_v(\x)=[W,P_v]^\top \phi(\x)$, where $P_v=[\p_1,\cdots,\p_V]\in\mathbb{R}^{d\times V}$ is the collection of virtual prototypes. The output probability after softmax operation is denoted as: $\bm a= \text{Softmax}\left([W,P_v]^\top \phi\left(\x\right)\right)= [a_1,\cdots,a_{|Y_0|+V}]$. We decouple the embedding module into two parts, \ie, $\phi(\x)=g(h(\x))$.

The final loss is defined as: $\mathcal{L}=\mathcal{L}_v+\mathcal{L}_f$, where
\begin{align} \label{eq:l1}
	\mathcal{L}_v(\x,y)=\underbrace{\ell (f_v(\x),y)}_{\mathcal{L}_1}	+&\gamma\underbrace{\,\ell\left(\text{Mask}(f_v(\x),y),\hat{y}\right)}_{\mathcal{L}_2}\\  \label{eq:l2}
	\mathcal{L}_f(\z)=	\underbrace{\ell (f_v(\z),\hat{y})}_{\mathcal{L}_3}+&\gamma\underbrace{\,\ell\left(\text{Mask}(f_v(\z),\hat{y}),\doublehat{y}\right)}_{\mathcal{L}_4} \,.
\end{align}
In Eq.~\ref{eq:l2}, $\z$ is the manifold mixup instance from two different classes, \ie, $\z = g\left[\lambda h(\x_i)+(1-\lambda)h(\x_j)\right], y_i\ne y_j$. In the following, we analyze the gradient using cross-entropy as the loss function $\ell(\cdot,\cdot)$.

Following the assumption in the main paper, in the analysis of gradients, we treat the classifier $[W,P_v]$ as a unified classifier, \ie, denote $[W,P_v]=[\w_1,\cdots,\w_{|Y_0|+V}]$ for ease of discussion. 

\subsection{Supplementary Analysis for the Main Paper}
{\vspace{4mm}}\noindent\bfname{Analyzing $\mathcal{L}_1$:}

The optimization target of $\mathcal{L}_1$ corresponds to:
\begin{align} \notag
	\mathcal  L_1=-\log a_y \,,
\end{align}
where the output probability is defined as:
\begin{align}
	a_i=\frac{\exp\left(\bm w_i^\top \phi(\x)\right)}{\sum_{j=1}^{\mid  Y_0\mid+ V }\exp\left({\bm w_j^\top}\phi(\x)\right)} \,.
\end{align}

Hence, we can obtain the negative gradient w.r.t. the embedding $\phi(\x)$:
\begin{align}\begin{split} \label{eq:l1phi}
		-\nabla_{\phi(\x)}{\mathcal L_1}&=\bm w_y -\sum_{i=1}^{\mid  Y_0\mid + V}a_i\bm w_{i}\,. 
	\end{split}
\end{align}
Eq.~\ref{eq:l1phi} indicates that optimizing $\mathcal{L}_1$ will push the embedding $\phi(\x)$ towards the direction of $\w_y$, and away from other prototypes. It is the classical loss function, which helps to acquire the classification ability and discriminability among known classes. 

We can also obtain the negative gradient w.r.t. the prototype $\w_i$:
\begin{align} \label{eq:l1wi}
	-\nabla_{\bm w_i}\mathcal L_1=\begin{cases}
		(1-a_i)\phi(\x), & i=y\\
		-a_i \phi(\x), & otherwise 
	\end{cases} \,.
\end{align}
Note that $a_i\in(0,1)$, which reflects the similarity between $\w_i$ and $\phi(\x)$. As a result, for the prototype weight from the ground truth label, \ie, $\w_y$, the smaller $a_y$ is, the larger the gradient norm is. For the prototype from non-target class, \ie, $\w_k, k\neq y$, the larger $a_k$ is, the larger the gradient norm is.
Eq.~\ref{eq:l1wi} means that optimizing $\mathcal{L}_1$ will push the class prototype of the target class, \ie, $\w_y$ towards the embedding of $\phi(\x)$, and push the non-target class prototypes away from it. The effect of Eq.~\ref{eq:l1wi} is consistent with that of Eq.~\ref{eq:l1phi}, which helps to classify $(\x,y)$ correctly.

{\vspace{4mm}}\noindent\bfname{Analyzing $\mathcal{L}_2$}

We denote the pseudo label assigned to $\x$ is $\hat{y}$, and $\mathcal{L}_2$ equals to:
\begin{align} \notag
	\mathcal L_2=-\log a_{\hat{y}} \,.
\end{align}
Note that when calculating $\mathcal{L}_2$, the output probability is masked out for the ground-truth class $y$, which yields:
\begin{align}
	a_i=\begin{cases}
		0, &i=y\\
		\frac{\exp\left(\bm w_i^\top \phi(\x)\right)}
		{\sum_{j=1}^{|Y_0|+V} \left(\exp\left({\bm w_j^\top}\phi(\x)\right) \right)-\exp\left({\bm w_y^\top}\phi(\x)\right) }  ,& otherwise\\
	\end{cases} \,.
\end{align}
Hence, we can obtain the negative gradient w.r.t. the embedding $\phi(\x)$:
\begin{align} \begin{split} \label{eq:l2phi}
		-\nabla_{\phi(\x)}{\mathcal L_2}&=\bm w_{\hat{y}} -\sum_{i=1}^{\mid  Y_0\mid +  V}a_i\bm w_{i}\,.
	\end{split}
\end{align}

Eq.~\ref{eq:l2phi} indicates that optimizing $\mathcal{L}_2$ will push the embedding $\phi(\x)$ towards the direction of $\w_{\hat{y}}$, and away from other prototypes. As a result, we reserve the embedding space for new classes explicitly by pushing other prototypes away to make the model growable and forward compatible.
Note that $a_y=0$ and the push effect will not influence the ground truth class $y$, \ie, optimizing Eq.~\ref{eq:l2phi} will not weaken the classification performance on known classes.

We can also obtain the negative gradient w.r.t. the prototype $\w_i$:
\begin{align} \label{eq:l2wi}
	-\nabla_{\bm w_i}\mathcal L_2=\begin{cases}
		(1-a_i)\phi(\x), & i=\hat{y}\\
		-a_i \phi(\x), & otherwise
	\end{cases} \,.
\end{align}
Eq.~\ref{eq:l2wi} means that optimizing $\mathcal{L}_2$ will push the class prototype of the target class, \ie, $\w_{\hat{y}}$ towards the embedding of $\phi(\x)$, and push the non-target class prototypes away from it. Note that the probability $a_y$ is 0 for the ground truth class, and optimizing $\mathcal{L}_2$ will not hurt the classification performance. 
The effect of Eq.~\ref{eq:l2wi} is consistent with that of Eq.~\ref{eq:l2phi}, which helps reserve the embedding space for class $\hat{y}$ explicitly without harming the classification performance.

{\vspace{4mm}}\noindent\bfname{Analyzing $\mathcal{L}_3$}

We denote the pseudo label assigned to $\z$ is $\hat{y}$, where $\z$ is the product of manifold mixup, \ie, $\z = g\left[\lambda h(\x_i)+(1-\lambda)h(\x_j)\right], y_i\ne y_j$. Similar to the assumption made in the main paper, we assume $g(\cdot)$ is identity function and $h(\x)=\phi(\x)$. We analyze the scenario when $g(\cdot)$ is not identity in Sec.~\ref{sec:notidentical}. $\mathcal{L}_3$ equals to:
\begin{align} \notag
	\mathcal L_3=-\log a_{\hat{y}} \,,
\end{align}
where output probability on class $m$ is defined as:
\begin{align}
	a_m=\frac{\exp\left(\w_m^\top \bm z\right)}{\sum_{k=1}^{\mid  Y_0\mid+ V }\exp\left({\w_k^\top}\bm z\right)} \,.
\end{align}
Hence, we can obtain the negative gradient w.r.t. the mixed embedding $\z$:
\begin{align} \label{eq:l3z}
	\begin{split}
		-\nabla_{\bm z}{\mathcal L_3}&=\bm w_{\hat{y}} -\sum_{k=1}^{\mid  Y_0\mid +  V}a_k\bm w_{k}\,.
	\end{split}
\end{align}
Eq.~\ref{eq:l3z} indicates that optimizing $\mathcal{L}_3$ will push the embedding $\z$ towards the direction of $\w_{\hat{y}}$, and away from other prototypes. Since $\z$ is a generated virtual instance, 
we reserve the embedding space for new classes explicitly by pushing other prototypes away to make the model provident and forward compatible.

We can also obtain the negative gradient w.r.t. the prototype $\w_k$:
\begin{align} \label{eq:l3wi}
	-\nabla_{\bm w_k}\mathcal L_3=\begin{cases}
		(1-a_k)\bm z, & k=\hat{y}\\
		-a_k \bm z, & otherwise
	\end{cases} \,.
\end{align}
Eq.~\ref{eq:l3wi} means that optimizing $\mathcal{L}_3$ will push the class prototype of the target class, \ie, $\w_{\hat{y}}$ towards the embedding of $\z$, and push the non-target class prototypes away from it. 
The effect of Eq.~\ref{eq:l3wi} is consistent with that of Eq.~\ref{eq:l3z}, which helps reserve the embedding space for class $\hat{y}$ explicitly by instance mixture.

Apart from the gradient w.r.t. the embedding $\z$, we also provide that for $\phi(\x_i)$ and $\phi(\x_j)$:
\begin{align} \label{eq:l3phi}
	\begin{split}
		-\nabla_{\phi({\x_i})}{\mathcal L_3}&=-\lambda \nabla_{\bm z}{\mathcal L_3}\\
		&=\lambda \left( \bm w_{\hat{y}} -\sum_{k=1}^{\mid  Y_0\mid +  V}a_k\bm w_{k} \right)\\
		-\nabla_{\phi({\x_j})}{\mathcal L_3}&=-(1-\lambda)\nabla_{\bm z}{\mathcal L_3} \\
		&= (1-\lambda) \left(\bm w_{\hat{y}} -\sum_{k=1}^{\mid  Y_0\mid +  V}a_k\bm w_{k} \right)\,,
	\end{split}
\end{align}
which indicates that the embedding of mixup components $\phi(\x_i)$ and $\phi(\x_j)$ will be pushed towards the prototype $\w_{\hat{y}}$, and away from other prototypes. Eq.~\ref{eq:l3phi} is similar to Eq.~\ref{eq:l2phi}, which reserves the embedding space for new classes and make the model forward compatible.

{\vspace{4mm}}\noindent\bfname{Analyzing $\mathcal{L}_4$}

The analysis of $\mathcal{L}_4$ is similar to that of $\mathcal{L}_3$. Assume the pseudo label assigned to the masked probability is $\doublehat{y}$, $\mathcal{L}_4$ is defined as:
\begin{align} \notag
	\mathcal L_4=-\log a_{\hathat{y}} \,,
\end{align}
where the output probability for class $m$ is defined as:
\begin{align}
	a_m=\begin{cases}
		0, &m={\hat{y}}\\
		\frac{\exp\left(\bm w_m^\top \bm z\right)}
		{\sum_{k=1}^{|Y_0|+V}\left(\exp\left({\bm w_k^\top}\bm z\right)\right) - \exp\left({\bm w_{\hat{y}}^\top}\bm z\right)},& otherwise\\ 
	\end{cases}\,.
\end{align}
Similarly, we have the negative gradient w.r.t. the manifold mixup product $\z$:
\begin{align} \begin{split}
		-\nabla_{\bm z}{\mathcal L_4}&=\bm w_{\hathat{y}} -\sum_{k=1}^{\mid Y_0\mid +  V}a_k\bm w_{k}\,.
	\end{split} 
\end{align}
The negative gradient w.r.t. the classifier weight yields:
\begin{align}
	-\nabla_{\bm w_k}\mathcal L_4=\begin{cases}
		(1-a_k)\bm z, & k=\doublehat{y}\\
		-a_k \bm z, & otherwise
	\end{cases} \,.
\end{align}
The negative gradient w.r.t. the mixup components yields:
\begin{align} \label{eq:l4phi}
	\begin{split}
		-\nabla_{\phi({\x_i})}{\mathcal L_4}&=-\lambda \nabla_{\bm z}{\mathcal L_4}\\
		&=\lambda \left( \bm w_{\hathat{y}} -\sum_{k=1}^{\mid  Y_0\mid +  V}a_k\bm w_{k} \right)\\
		-\nabla_{\phi({\x_j})}{\mathcal L_4}&=-(1-\lambda)\nabla_{\bm z}{\mathcal L_4} \\
		&= (1-\lambda) \left(\bm w_{\hathat{y}} -\sum_{k=1}^{\mid  Y_0\mid +  V}a_k\bm w_{k} \right)\,,
	\end{split}
\end{align}

To summarize, when considering $\mathcal{L}_4$, the final loss function becomes symmetric. The effect of  $\mathcal{L}_4$ is similar to that of  $\mathcal{L}_1$. Since $\mathcal{L}_2$ and $\mathcal{L}_3$ both reserve the embedding space for new classes and squeeze the space of known ones, we seek to trade-off between the squeeze process and avoid over-squeezing with such symmetric loss form. The ablations in the main paper validate the effectiveness of such regularization.

\subsection{When $g(\cdot)$ is not Identity} \label{sec:notidentical}

In the former part and main paper, we analyze the gradients of $\mathcal{L}_3$ and $\mathcal{L}_4$ by assuming $g(\cdot)$ is the identity function. In this section, we analyze a more general scenario where $g(\cdot)$ is not identity. We first analyze the liner situation and then analyze the nonlinear situation. Note that the gradients of $\mathcal{L}_3$ and $\mathcal{L}_4$ are similar, and we only give the gradients of $\mathcal{L}_3$ as an example.

\subsubsection{$g(\cdot)$ is Linear}
Following the former analysis, if $g(\cdot)$ is a linear layer, we can parameterize it as $\mathbf V$, and we have $\phi(\x)=\mathbf V^\top h(\x)$. Denote the mixed instance at middle layer as $\bm b=\lambda h(\x_i)+(1-\lambda)h(\x_j)$, then $\z=g(\bm b)=\mathbf V^\top \bm b$ indicates the embedding of mixed instance. Hence, the output probability for mixed instance on class $m$ is defined as:
\begin{align} \notag
	a_{m}=\frac{\exp \left(\bm {w}_{m}^{\top}\mathbf V^\top \bm {b}\right)}{\sum_{k=1}^{\left|{Y}_{0}\right|+V} \exp \left(\bm {w}_{k}^{\top}\mathbf V^\top \bm {b}\right)} \,.
\end{align}
We can obtain the negative gradient w.r.t. the final embedding of mixed instance $g(\bm b)$:
\begin{align}
	\begin{split}
		-\nabla_{g(\bm {b})} \mathcal{L}_{3} &=\bm {w}_{\hat{y}}-\sum_{k=1}^{\left|{Y}_{0}\right|+V} a_{k} \bm {w}_{k} \,,
	\end{split}
\end{align}
which is same to Eq.~\ref{eq:l3z}. The gradients indicate that optimizing $\mathcal{L}_3$ will
reserve the embedding space for class $\hat{y}$ by moving $\mathbf V^\top \bm b$ towards $\w_{\hat{y}}$, and away from other prototypes. We can further obtain the negative gradients w.r.t. $\bm b$:
\begin{align} \label{eq:linearb}
	\begin{split}
		-\nabla_{\bm {b}}\mathcal L_3&=-\mathbf V \nabla_{g(\bm b)}\mathcal L_3\\
		&=\mathbf V\left(\bm {w}_{\hat{y}}-\sum_{k=1}^{\left|{Y}_{0}\right|+V} a_{k} \bm {w}_{k} \right) \,,
	\end{split}
\end{align}
which indicates that the pushing effect also works in the middle layer (\ie, the output layer of $h(\x)$), encouraging the feature reserving in the same direction as the last layer.
Eq.~\ref{eq:linearb} verifies that the reserving process works from shallow to deep, which makes forward compatibility maintained holistically.
Then we get the negative gradients w.r.t. the mixup components $h(\x_i)$ and $h(\x_j)$:
\begin{align}\notag
	-\nabla_{h(\mathbf x_i)}\mathcal L_3=\lambda \mathbf V\left(\bm {w}_{\hat{y}}-\sum_{k=1}^{\left|{Y}_{0}\right|+V} a_{k} \bm {w}_{k} \right)
\end{align}
\begin{align}\notag
	-\nabla_{h(\mathbf x_j)}\mathcal L_3=(1-\lambda) \mathbf V\left(\bm {w}_{\hat{y}}-\sum_{k=1}^{\left|{Y}_{0}\right|+V} a_{k} \bm {w}_{k} \right) \,.
\end{align}
The conclusions are consistent with Eq.~\ref{eq:l3phi}, which only adds an extra term in the gradient direction, and we can infer that even $g(\cdot)$ is a linear classifier, the forward compatibility is still maintained with $\mathcal{L}_3$.

\subsubsection{$g(\cdot)$ is Nonlinear}
Under a more common scenario, we assume  $g(\cdot)$ is nonlinear (\eg, by a neural network block), and we have:
\begin{align} \notag
	a_{m}=\frac{\exp \left(\bm {w}_{m}^{\top}g(\bm {b})\right)}{\sum_{k=1}^{\left|{Y}_{0}\right|+V} \exp \left(\bm {w}_{k}^{\top}g (\bm {b})\right)} \,.
\end{align}
Similarly, we have the negative gradients w.r.t. the embedding $g(\bm {b})$:
\begin{align}
	\begin{split}
		-\nabla_{g(\bm {b})} \mathcal{L}_{3} &=\bm {w}_{\hat{y}}-\sum_{k=1}^{\left|{Y}_{0}\right|+V} a_{k} \bm {w}_{k} \,.
	\end{split} 
\end{align}
The negative gradient w.r.t. the mixed instance $\bm b$ can be obtained via:
\begin{align}
	\begin{split}
		-\nabla_{\bm {b}}\mathcal L_3=&-(\nabla_{\bm b}\bm J)^\top\nabla_{g(\bm b)}\mathcal L_3\\
		=&(\nabla_{\bm b}\bm J)^\top\left(\bm {w}_{\hat{y}}-\sum_{k=1}^{\left|{Y}_{0}\right|+V} a_{k} \bm {w}_{k}\right) \,,
	\end{split}
\end{align}
where $\nabla_{\bm b}\bm J$ is the Jacobian matrix of $g(\bm b)$ w.r.t. $\bm b$. Suppose that $\left(\nabla_{\bm b}\bm J\right)^\top\w_{\hat{y}}$ are pointing to similar directions (\eg, with a high cosine similarity), then the conclusion above still holds.
Similarly, we have the negative gradient w.r.t. the mixup components $h(\x_i)$ and $h(\x_j)$:
\begin{align}\notag
	-\nabla_{h(\mathbf x_i)}\mathcal L_3=\lambda (\nabla_{\bm b}\bm J)^\top\left(\bm {w}_{\hat{y}}-\sum_{k=1}^{\left|{Y}_{0}\right|+V} a_{k} \bm {w}_{k} \right)
\end{align}
\begin{align} \notag
	-\nabla_{h(\mathbf x_j)}\mathcal L_3=(1-\lambda)(\nabla_{\bm b}\bm J)^\top\left(\bm {w}_{\hat{y}}-\sum_{k=1}^{\left|{Y}_{0}\right|+V} a_{k} \bm {w}_{k}\right) \,.
\end{align}
The conclusions are consistent with Eq.~\ref{eq:l3phi}, and we can infer that even $g(\cdot)$ is not a linear classifier, the forward compatibility is still maintained with $\mathcal{L}_3$.
To conclude, $\mathcal{L}_3$ enhances the forward compatibility holistically.

\section{Degradation Form of \name}\label{sec:degradation} 

In this section, we give another inference form of \mame, which adopts another form of assumption and can be seen as a degradation form of \mame. 
With a bit of redundancy, we start from the law of total probability:
\begin{align} \label{eq:unfold2}
	\begin{split}
		p\left(y_i| {\phi(\x)}\right)&=p(\bm w_{i}| {\phi(\x)})\\
		&=\sum_{\p_v \in  P_v}p(\bm w_i| \p_{v},{\phi(\x)})p(\p_{v}| {\phi(\x)}) \,,
	\end{split}
\end{align}
where $
p(\p_{v}|{\phi(\x)})=\frac{\exp\left({\p_{v}}^\top {\phi(\x)}\right)}{\sum_{\p_{v} \in P_v}\exp\left({\p_{v}}^\top {\phi(\x)}\right)} 
$. Eq.~\ref{eq:unfold2} implies that we can consider the possible influence of all informative virtual prototypes to get the final prediction.
We still assume $p\left(\phi(\mathbf{x})| \bm{w}_{i},\bm{p}_{v}\right)=\eta \mathcal N(\phi(\x)| \bm w_i,\Sigma)+(1-\eta)\mathcal N(\phi(\mathbf x)| \bm p_v,\Sigma)$, which follows a Gaussian mixture distribution.
According to Bayes' Theorem, we have:
\begin{align} \notag
	p(\bm w_i|{\p_v,\phi(\x)} )=\frac{p({\phi(\x)}| \bm w_i, \p_v) p(\bm w_i| \p_v)}
	{\sum_{j=1}^{|\mathcal{Y}_b|}p({\phi(\x)}| \bm w_j, \p_v) p(\bm w_j| \p_v)} \,,
\end{align}
where $|\mathcal{Y}_b|$ is the number of classes seen before. In the main paper, we argue that $p(\w_i|\p_v)$ reflects the similarity between $\w_i$ and $\p_v$, and assume it follows a Gaussian distribution. However, we can also treat it as the class prior given virtual class, which can be discarded by assuming all classes follow a uniform distribution in few-shot class-incremental learning. Hence, We have:
\begin{align} \label{eq:inference2}
	\begin{split}
		p(\bm w_i| & \p_v,{\phi(\x)})=\frac{p({\phi(\x)}| \bm w_i, \bm p_v)}{\sum_{j=1}^{|\mathcal{Y}_b|}p({\phi(\x)}| \bm w_j, \bm p_v) }\\
		&=\frac{\eta \bm n(\bm w_i)+(1-\eta) \bm n(\bm p_v)} {\eta\sum_{j=1}^{\left|\mathcal Y_b\right|} \bm n(\bm w_j)+ (1-\eta)|\mathcal{Y}_b|\bm n(\bm p_v)} \,,
	\end{split}
\end{align}
where $\bm n(\bm w)=\exp \left(\left(\Sigma^{-1} \bm w\right)^{\top} \phi(\mathbf{x})-\frac{1}{2} \bm w^{\top} \Sigma^{-1} \bm w\right)$. When $\w$ and $\p$ are normalized, $\Sigma=I, \eta=1$, Eq.~\ref{eq:inference2} degrades into:
\begin{align} \notag
	p(\bm w_i|  \p_v,{\phi(\x)})&=\frac{\exp \left(\bm \w_i^\top {\phi(\x)}\right)} {\sum_{j=1}^{|\mathcal{Y}_b|} \exp \left(\bm \w_j^\top {\phi(\x)}\right)}\,,
\end{align}
which means the probability is irrelevant to the virtual prototype $\p_v$. Hence, Eq.~\ref{eq:unfold2} turns into:
\begin{align}
	\begin{split}
		p\left(y_i| {\phi(\x)}\right)&=\sum_{\p_v \in  P_v}p(\bm w_i| \p_{v},{\phi(\x)})p(\p_{v}| {\phi(\x)}) \\
		&= \frac{\exp \left(\bm \w_i^\top {\phi(\x)}\right)} {\sum_{j=1}^{|\mathcal{Y}_b|} \exp \left(\bm \w_j^\top {\phi(\x)}\right)}  \,.
	\end{split}
\end{align}
It degrades into ProtoNet (as discussed in Section~3.2 of the main paper), where we only consider the influence of known class prototypes and ignore the possible influence of virtual prototypes. However, in our ablation study (c.f. Section~5.3 of the main paper), we find that when training the model with the same loss function, the inference performance of our \name is better than ProtoNet. It validates that the classification ability is encoded into these virtual prototypes, which can differentiate among all known classes and help build a stronger classifier.

\begin{table*}[t]
	\centering{
		\caption{Comparison with the state-of-the-art on CIFAR100 dataset. We report the results of compared methods from~\cite{tao2020few} and~\cite{zhang2021few}. \textbf{\name outperforms the runner-up method by 2.96\% in terms of the last accuracy, by 1.43\% in terms of  the performance decay.}  }\label{tab:cifar}
		\resizebox{\textwidth}{!}{
			\begin{tabular}{llllllllllll}
				\toprule
				\multicolumn{1}{c}{\multirow{2}{*}{Method}} & \multicolumn{9}{c}{Accuracy in each session (\%) $\uparrow$} & \multirow{2}{*}{PD $\downarrow$} & {Our relative
				} \\ \cmidrule{2-10}
				\multicolumn{1}{c}{} & 0   & 1      & 2      & 3    & 4     & 5  & 6     & 7      & 8     &     &  improvement      \\ \midrule
				Finetune                & 64.10   & 39.61      & 15.37      & 9.80   & 6.67     & 3.80  & 3.70    & 3.14      & 2.65     & 61.45   & \bf +38.95     \\
				Pre-Allocated RPC~\cite{pernici2021class}& 64.50& 54.93& 45.54& 30.45& 17.35& 14.31& 10.58& 8.17& 5.14 &59.36 & \bf +36.86\\

				iCaRL~\cite{rebuffi2017icarl}       & 64.10   & 53.28      & 41.69      & 34.13   & 27.93     & 25.06  & 20.41   & 15.48      &13.73    & 50.37  & \bf +27.87  \\
				EEIL~\cite{castro2018end}         & 64.10   & 53.11     & 43.71     & 35.15   & 28.96     & 24.98  & 21.01    & 17.26     & 15.85    & 48.25  & \bf +25.75    \\
				Rebalancing~\cite{hou2019learning}           & 64.10   & 53.05     & 43.96      & 36.97   & 31.61     & 26.73  & 21.23   & 16.78    & 13.54     &50.56  & \bf +28.06   \\
				TOPIC~\cite{tao2020few}                & 64.10   & 55.88     & 47.07      & 45.16   & 40.11   & 36.38 & 33.96   & 31.55      & 29.37     & 34.73   & \bf +12.23      \\
				Decoupled-NegCosine~\cite{liu2020negative}&
				74.36& 68.23& 62.84& 59.24& 55.32& 52.88& 50.86& 48.98& 46.66&27.70 & \bf +5.20 \\
				Decoupled-Cosine~\cite{vinyals2016matching}  & 74.55   & 67.43      & 63.63      & 59.55  & 56.11    & 53.80  & 51.68   & 49.67     & 47.68     & 26.87  & \bf +4.37     \\
				Decoupled-DeepEMD~\cite{zhang2020deepemd}    & 69.75   & 65.06     & 61.20     & 57.21  & 53.88    & 51.40  & 48.80  & 46.84     & 44.41     & 25.34   & \bf +2.84    \\
				CEC~\cite{zhang2021few}                    & 73.07   & 68.88     & 65.26    & 61.19  & 58.09   &55.57  & 53.22   & 51.34     & 49.14   & 23.93   & \bf +1.43    \\
				\midrule
				
				\name        & \bf 74.60   & \bf 72.09    & \bf67.56   & \bf63.52 & \bf61.38  &\bf58.36 & \bf56.28   &\bf  54.24     & \bf52.10   & \bf22.50   &   \\

				\bottomrule
			\end{tabular}
	}}
\end{table*}

\begin{table*}[t]
	\centering{
		\caption{Comparison with the state-of-the-art on \textit{mini}ImageNet dataset. We report the results of compared methods from~\cite{tao2020few} and~\cite{zhang2021few}. \textbf{\name outperforms the runner-up method by 2.86\% in terms of the last accuracy, by 2.30\% in terms of  the performance decay.} }\label{tab:mini}
		\resizebox{\textwidth}{!}{
			
			\begin{tabular}{llllllllllll}
				\toprule
				\multicolumn{1}{c}{\multirow{2}{*}{Method}} & \multicolumn{9}{c}{Accuracy in each session (\%) $\uparrow$} & \multirow{2}{*}{PD $\downarrow$} & {Our relative
				} \\ \cmidrule{2-10}
				\multicolumn{1}{c}{} & 0   & 1      & 2      & 3    & 4     & 5  & 6     & 7      & 8     &     &  improvement      \\ \midrule
				Finetune                & 61.31  & 27.22      & 16.37     & 6.08   & 2.54     & 1.56  & 1.93    & 2.60      & 1.40     & 59.91   & \bf +37.84    \\
				Pre-Allocated RPC~\cite{pernici2021class} & 61.25&
				31.93& 18.92& 13.90& 14.37& 15.57& 16.15& 12.33& 12.28 & 48.97 &\bf +26.90\\
				
				iCaRL~\cite{rebuffi2017icarl}       & 61.31   & 46.32     & 42.94      & 37.63   & 30.49     & 24.00  & 20.89   & 18.80      &17.21  & 44.10  & \bf +22.03 \\
				EEIL~\cite{castro2018end}         & 61.31   & 46.58     & 44.00    & 37.29   & 33.14    & 27.12  & 24.10    & 21.57     & 19.58    & 41.73 & \bf +19.66  \\
				Rebalancing~\cite{hou2019learning}           & 61.31   & 47.80     & 39.31      & 31.91  & 25.68     & 21.35  & 18.67   & 17.24    & 14.17     &47.14  & \bf +25.07    \\
				TOPIC~\cite{tao2020few}                & 61.31 & 50.09    & 45.17      & 41.16   & 37.48   & 35.52 & 32.19  & 29.46    & 24.42     & 36.89   & \bf +14.82      \\
				Decoupled-NegCosine~\cite{liu2020negative}&
				71.68& 66.64&62.57& 58.82& 55.91& 52.88& 49.41& 47.50& 45.81&25.87 & \bf +3.80 \\
				
				Decoupled-Cosine~\cite{vinyals2016matching}  & 70.37  & 65.45      & 61.41     & 58.00  & 54.81   & 51.89  & 49.10   & 47.27     & 45.63    & 24.74 & \bf +2.67   \\
				Decoupled-DeepEMD~\cite{zhang2020deepemd}    & 69.77   & 64.59     & 60.21    & 56.63  & 53.16   & 50.13  & 47.79  & 45.42     & 43.41    & 26.36   & \bf +4.29   \\
				CEC~\cite{zhang2021few}                    & 72.00   & 66.83     & 62.97   & 59.43  & 56.70   &53.73  & 51.19   & 49.24     & 47.63   & 24.37   & \bf +2.30     \\
				\midrule
				
				\name              & \bf 72.56   & \bf 69.63     & \bf66.38    & \bf62.77  & \bf60.6  &\bf57.33  & \bf54.34  &\bf 52.16     & \bf 50.49   & \bf22.07   &   \\
				\bottomrule
				
			\end{tabular}
	}}
\end{table*}

\section{Introduction about Compared Methods }\label{sec:compared}

In this section, we give a detailed introduction about the compared methods adopted in the main paper. They are listed as:

\begin{itemize}

	\item{ \bfname{Finetune}}: when facing the few-shot incremental session, it simply optimizes the cross-entropy over these few-shot items. It easily suffers catastrophic forgetting.

	\item{ \bfname{iCaRL}~\cite{rebuffi2017icarl}}: when training an incremental new task, it combines cross-entropy loss with knowledge distillation loss together. The knowledge distillation part can help the model maintain discrimination ability over former learned knowledge.
	
	\item {\bfname{	Pre-Allocated RPC~\cite{pernici2021class}}}: it is a class-incremental learning method based on data rehearsal. It first pre-allocates all the classifiers on a regular polytope and then optimizes the embedding of new classes to fit these pre-allocated classifiers. However, since FSCIL tasks do not save exemplars for rehearsal, we can only use the few-shot dataset to optimize the embedding.
	
	\item{ \bfname{EEIL}~\cite{castro2018end}}: considers an extra balanced fine-tuning process over iCaRL, which uses a balanced dataset to finetune the model and alleviate bias.
	
	\item{ \bfname{Rebalancing}~\cite{hou2019learning}}: uses cosine normalization, feature-wise knowledge distillation and contrastive learning to augment the model and resist catastrophic forgetting.
	
	\item{ \bfname{TOPIC}~\cite{tao2020few}}: tailors few-shot class-incremental learning task with neural gas network. It preserves the topology of the feature	manifold formed by different classes.
	
	\item{ \bfname{Decoupled-DeepEMD}~\cite{zhang2020deepemd}}: decouples the training process of embedding and classifier. After the embedding training process of the base session, it replaces the classifier of each class with the mean embedding of this class. When learning a new session, the same classifier replacement process is adopted for every new class.
	It adopts a DeepEMD distance~\cite{zhang2020deepemd} calculation between classifier and incoming queries during inference.
	
	\item{ \bfname{Decoupled-Cosine}~\cite{vinyals2016matching}}:  Similar to Decoupled-DeepEMD,  it decouples the training process of embedding and classifier.
	It adopts a cosine distance~\cite{vinyals2016matching} calculation  during inference.
	
	\item{ \bfname{Decoupled-NegCosine}~\cite{liu2020negative}}: Similar to Decoupled-Cosine,  it decouples the training process of embedding and classifier.
	The difference between it and Decoupled-Cosine is that it uses a negative margin softmax function during model pretraining.
	It adopts a cosine distance~\cite{vinyals2016matching} calculation  during inference.
	
	\item{ \bfname{CEC}~\cite{zhang2021few}}: trains an extra graph model during base session with pseudo-incremental learning sampling. The graph model learns to adapt the embeddings of old class prototypes and new class prototypes, and such ability is generalizable to the incremental learning process.

\end{itemize}

Note that iCaRL, EEIL, Pre-Allocated RPC, and Rebalancing are traditional class-incremental algorithms. Our empirical experiments in the main paper indicate that these classical class-incremental methods are unsuitable for few-shot class-incremental learning scenarios. For other SOTA methods of FSCIL, our proposed \name consistently outperforms them by vast performance measures.

\noindent\textbf{Discussion about related compatible training methods:} Some other works aim to build a compact embedding space~\cite{zhou2021co}, which can be seen as enhancing forward compatibility implicitly. For example, \cite{zhou2021learning} seeks to detect new classes by learning placeholders, \cite{liu2016large} utilizes the embedding with large margin between classes, \cite{ranasinghe2021orthogonal} encourages class-wise orthogonality for more compact embedding. These works, however, have a different goal from ours. Their ultimate goal is to obtain a compact embedding, which facilitates the one-stage or in-domain performance in anomaly detection or classification. By contrast, our training scheme aims to build an embedding that enhances performance in the future. In other words, we propose forward compatibility to tailor the characteristics of FSCIL with multiple tasks. Besides, there are other differences between ours and~\cite{zhou2021learning}. For example, we use a symmetric loss to balance the learning and reserving process, which is consistent with forward compatibility and proven efficient. Besides, we use the class prototype (average mean) as the classifier, which benefits the embedding learning process in FSCIL. Lastly, the former learned virtual prototypes are utilized during inference to boost forward compatibility, instead of dropped directly. There are other methods addressing virtual classes, \eg,~\cite{chou2020adaptive}. However, the setting in~\cite{chou2020adaptive} is different from ours, where extra semantic information are available to synthesis new classes directly.

Pre-Allocated RPC and Decoupled-NegCosine can be viewed as encouraging forward compatibility. The former pre-assigns the classifier for new classes but lacks the ability for classifier matching  in FSCIL process. The latter considers preventing new class embedding from being harmed by using a negative margin. 
However, they are validated ineffective in the FSCIL setting, verifying the effectiveness of our prospective training paradigm.

\section{Detailed Incremental Performance }\label{sec:incrementalperformance}
In the main paper, we show the incremental performance on benchmark datasets, and report the detailed accuracy of CUB200. We report the incremental performance on the other benchmark datasets, \ie, CIFAR100 and \textit{mini}ImageNet in Table~\ref{tab:cifar} and Table~\ref{tab:mini}. 
We can infer that our proposed \name has the better top-1 accuracy and lower performance decay, indicating \name forgets lower than other state-of-the-art methods. These conclusions are consistent with the main paper, verifying the best performance of \mame.

\begin{algorithm}[t]
	\small
	\caption{\small Forward Compatible Training for FSCIL }
	\label{alg1}
	{\bf Input}: Base dataset: $\D^0$, Virtual class number: $V$;\\
	{\bf Output}: $W,P_v$, ${\phi}(\cdot)$; 
	\begin{algorithmic}[1]{
			\STATE Randomly initialize $W,P_v$, ${\phi}(\cdot)$;
			\REPEAT
			\STATE Get a mini-batch of training instances: $\{(\x_i,y_i)\}_{i=1}^{n}$;
			\STATE Calculate the virtual loss $\mathcal{L}_v$;
			\STATE Randomly shuffle the dataset, denoted as $\{(\x_j,y_j)\}_{j=1}^{n}$;
			\STATE Mask out same class instances in $(\x_i,\x_j) $;
			\STATE Calculate the forecasting loss $\mathcal{L}_f$;
			\STATE Get the total loss $\mathcal{L}=\mathcal{L}_v+\mathcal{L}_f$; 
			\STATE Obtain derivative and update the model;
			\UNTIL reaches predefined epoches
		}
	\end{algorithmic}
\end{algorithm}

\end{document}